\documentclass[
    aps,
    prx,
    reprint,
    superscriptaddress,
    letterpaper,
    twocolumn,
    longbibliography,
    floatfix
]{revtex4-2}
\usepackage[utf8]{inputenc}
\usepackage[T1]{fontenc}
\usepackage[pass]{geometry}
\usepackage{float}
\usepackage{graphicx}
\usepackage{bm}
\usepackage{braket}
\usepackage{xcolor}
\usepackage{amsmath,amssymb,amsthm,amsfonts}
\usepackage{dirtytalk}
\usepackage{url}
\usepackage{bbold}
\usepackage{realboxes}
\usepackage{bibunits}
\defaultbibliographystyle{apsrev4-2}
\defaultbibliography{main}

\usepackage{enumitem}
\usepackage[toc,page,titletoc]{appendix}

\usepackage{tocloft}
\definecolor{dark-red}{rgb}{0.4,0.15,0.15}
\definecolor{dark-blue}{rgb}{0.15,0.15,0.4}
\definecolor{medium-blue}{rgb}{0,0,0.5}
\usepackage[hypertexnames=false]{hyperref}
\usepackage[all]{hypcap}
\hypersetup{
colorlinks, linkcolor={dark-red},
citecolor={dark-blue}, urlcolor={medium-blue}
}
\usepackage{cleveref}

\usepackage{algpseudocode}
\usepackage{algorithm}                
\usepackage{algorithmicx,setspace}    

\newcommand{\Vin}{V^{\text{in}}}
\newcommand{\Vout}{V^{\text{out}}}

\newcommand{\cmmnt}[1]{}

\usepackage[autolanguage]{numprint}

\usepackage{listings}
\usepackage{xcolor}

\definecolor{bkgd}{RGB}{240,242,246}
\definecolor{ceruleanblue}{rgb}{0.16, 0.32, 0.75}
\definecolor{orange-red}{rgb}{1.0, 0.27, 0.0}
\definecolor{anotherblue}{RGB}{37,92,243}
\definecolor{blackblue}{RGB}{46,60,85}
\definecolor{goldyellow}{RGB}{199,146,12}
\lstdefinestyle{altstyle2}{
backgroundcolor=\color{bkgd},
basicstyle=\ttfamily\footnotesize\color{blackblue},
breakatwhitespace=false,
breaklines=true,
captionpos=b,
commentstyle=\color{goldyellow},
keepspaces=true,
keywordstyle=\color{orange-red},
language=Python,
numbersep=5pt,
numberstyle=\tiny\color{ceruleanblue},
showspaces=false,
showstringspaces=false,
showtabs=false,
stringstyle=\color{anotherblue},
tabsize=2
}
\begin{document}
\title{An efficient probabilistic hardware architecture for diffusion-like models}

\author{Andraž Jelinčič}
\thanks{These authors contributed equally to this work.}
\affiliation{Extropic Corp.}

\author{Owen Lockwood}
\affiliation{Extropic Corp.}

\author{Akhil Garlapati}
\affiliation{Extropic Corp.}

\author{Peter Schillinger}
\affiliation{Extropic Corp.}

\author{Isaac L. Chuang}
	\affiliation{Department of Electrical Engineering and Computer Science,
                    Massachusetts Institute of Technology}

\author{Guillaume Verdon}
\affiliation{Extropic Corp.}

\author{Trevor McCourt$\,^{*,\,}$}
\email[Corresponding author: ]{trevor@extropic.ai}
\affiliation{Extropic Corp.}

\date{\today}

\begin{abstract}

The proliferation of probabilistic AI has prompted proposals for specialized stochastic computers. Despite promising efficiency gains, these proposals have failed to gain traction because they rely on fundamentally limited modeling techniques and exotic, unscalable hardware. In this work, we address these shortcomings by proposing an all-transistor probabilistic computer that implements powerful denoising models at the hardware level. A system-level analysis indicates that devices based on our architecture could achieve performance parity with GPUs on a simple image benchmark using approximately \numprint{10000} times less energy.

\end{abstract}
\maketitle
\begin{bibunit}
\nocite{apsrev42Control}

The unprecedented recent investment in large-scale AI systems will soon put a strain on the world's energy infrastructure. Every year, U.S. firms spend an amount larger than the inflation-adjusted cost of the Apollo program on AI-focused data centers~\cite{ai_spend, stine2009manhattan}. By 2030, these data centers could consume around 10\% of all of the energy produced in the U.S.~\cite{aljbour2024powering}. At the same time, model sizes, dataset sizes, and deployment footprints continue to grow rapidly.

Existing AI systems based on autoregressive large language models (LLMs) are valuable tools in white-collar fields~\cite{alphacode, gptbar, nori2023capabilities, work_1, work_2, work_3}, and are being adopted by consumers faster than the internet~\cite{bick2024rapid}. However, LLMs were architected specifically for GPUs~\cite{vaswani2017attention}, hardware originally intended for graphics, whose suitability for machine learning was discovered accidentally decades later~\cite{ng_gpu,doc_gpu}. Had a different style of hardware been popular in the last few decades, AI algorithms would likely have evolved in a different direction, and possibly a more energy-efficient one. This interplay between algorithm research and hardware availability is known as the ``hardware lottery''~\cite{hooker2021hardware}, and it entrenches hardware–algorithm pairings that may be far from optimal.

Therefore, prudent planning calls for systematic exploration of other types of AI systems in search of energy-efficient alternatives. Active efforts include mixed-signal compute-in-memory accelerators~\cite{ibm_analog}, photonic neural networks~\cite{mit_photonic}, and neuromorphic processors that emulate biological spiking~\cite{spinn2, lohi2}. These approaches explicitly confront the fact that computation is a physical process and that respecting physical constraints can lead to qualitatively different designs.

Probabilistic computing is an attractive approach because it can connect directly to AI at the system level via Energy-Based Models (EBMs). EBMs are a well-established model class in contemporary deep learning and have been competitive with the state of the art in tasks like image generation and robotic path planning~\cite{song2019generative, janner2022planning}. Using probabilistic hardware to accelerate EBMs falls under the broad umbrella of \emph{thermodynamic computing}~\cite{conte2019thermodynamic}, in which computation is expressed in terms of stochastic dynamics on energy landscapes.

\begin{figure}[t]
\includegraphics[width=\linewidth]{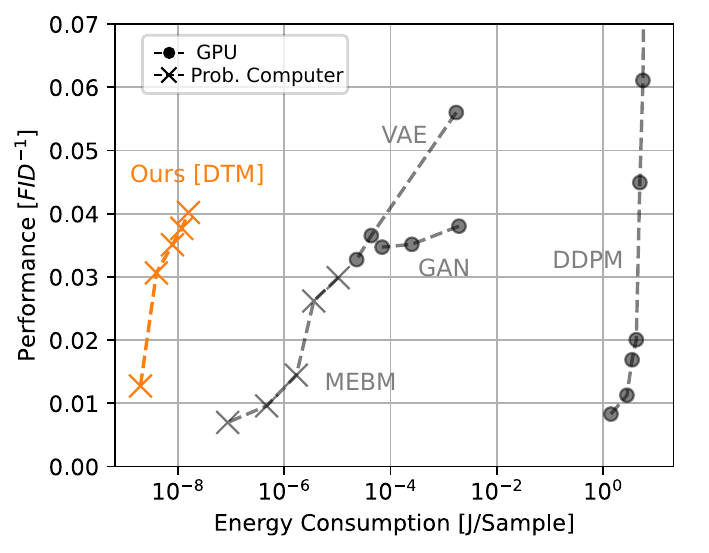}
\caption{\textbf{Leveraging CMOS probabilistic hardware in ultra-efficient AI systems.} The central result of this article: an all-transistor probabilistic computer running a denoising thermodynamic model (DTM) could match GPU performance on a simple modeling benchmark while using about $\numprint{10000} \times$ less energy. All models are trained on binarized Fashion-MNIST~\cite{xiao2017fashion} and evaluated with Fréchet Inception Distance (FID)~\cite{heusel2017ttur}. DTM variants are of increasing depth, chaining 2–8 sequential Energy-Based Models (EBMs). GPU baselines cover single-step VAE~\cite{kingma2022autoencodingvariationalbayes} and GAN~\cite{goodfellow2014generative}, plus DDPM~\cite{diff_orig} at varying numbers of steps. We also compare DTM to a monolithic EBM across multiple mixing-time limits. The horizontal axis shows the energy needed for generating a single new image using the trained model (inference).}
\label{fig:perf}
\end{figure}

Hardware implementations of EBMs work with special model families that adhere to physical constraints such as locality, sparsity, and limited connection density. Thanks to these constraints, probabilistic computers can utilize specialized stochastic circuitry to efficiently and quickly produce samples from a Boltzmann distribution~\cite{singh2024cmos}. Depending on the precise kind of hardware being used, this sampling may occur as part of the natural dynamics of the device~\cite{pratt_sc, pratt_sc_2,dwave, dutta_intrinsic, dutta_early} or may be orchestrated using an algorithm like Gibbs sampling~\cite{niazi2024training, kerem_factorization, kerem_learning, kerem_sparse}. 

Past attempts at EBM accelerators have suffered from issues at both the architectural and hardware levels. All previous proposals used EBMs as monolithic models of data distributions, which is known to be challenging to scale~\cite{du2019implicit}. Additionally, existing devices have relied on exotic components such as magnetic tunnel junctions as sources of intense thermal noise for random-number generation (RNG)~\cite{lee2025correlation, horodynski2025stochastic, kerem_learning}. These exotic components have not yet been tightly integrated with transistors in commercial CMOS processes and do not currently constitute a scalable solution~\cite{abeed_mtj, abeed_mtj_2, drob_mtj}.

In this work, we address these issues and propose a commercially viable probabilistic computing system that is designed for thermodynamic hardware from the outset. At the modeling level, we introduce Denoising Thermodynamic Models (DTMs), which repurpose hardware EBMs as denoising steps rather than monolithic models of the data distribution, thereby mitigating the mixing–expressivity tradeoff that plagues monolithic EBMs. At the architectural level, we propose the Denoising Thermodynamic Computer Architecture (DTCA), which tightly integrates DTMs into probabilistic hardware built from sparse, locally connected Boltzmann machines and an all-transistor RNG. Finally, we perform a system-level energy analysis, using measurements from an experimental thermodynamic RNG chip together with physical models and simulations, and show that a device based on this architecture could match the performance of GPU-based generative models on binarized Fashion-MNIST (as measured by Fréchet Inception Distance) while using approximately \numprint{10000} times less energy per generated sample.

The remainder of this article will substantiate the results presented in Fig.~\ref{fig:perf}, which are based on a combination of measurements from real circuits, physical models, and simulations. To begin, we introduce a fundamental compromise inherent in using EBMs as standalone models of data, which we refer to as the \emph{mixing–expressivity tradeoff}. We then discuss how this compromise can be avoided by wielding EBMs as part of a denoising process rather than monolithically. Next, we outline how to build a hardware system using DTMs to implement this denoising process at a very low level. Then, we study simulations of this hardware system, further justifying the results shown in Fig.~\ref{fig:perf} and highlighting some of the practical merits of DTMs compared to existing approaches. Finally, we conclude by discussing how the capabilities of probabilistic accelerators for machine learning may be scaled by merging them with traditional neural networks (NNs). We demonstrate that such a hybrid model composed of a DTM and a neural network can achieve performance parity with a purely NN-based model on the CIFAR-10 benchmark while using a ten times smaller neural network component (both in terms of the number of parameters and FLOPs) and thus consuming roughly an order of magnitude less energy in total.

\section{The Challenge with EBMs}

The fundamental problem of machine learning is inferring the probability distribution that underlies some data~\cite{mlsci, ghahramani2015probabilistic}. An early approach~\cite{hinton1984boltzmann} to this was to use a monolithic EBM (MEBM) to fit a data distribution directly by shaping a parameterized energy function $\mathcal{E}$:
\begin{equation}\label{eq:ebm}
P(x; \theta) \propto e^{- \mathcal{E}(x, \theta)},
\end{equation}
where $x$ is a random variable representing the data and $\theta$ represents the parameters of the EBM.

\begin{figure}
\includegraphics[width=0.75\linewidth]{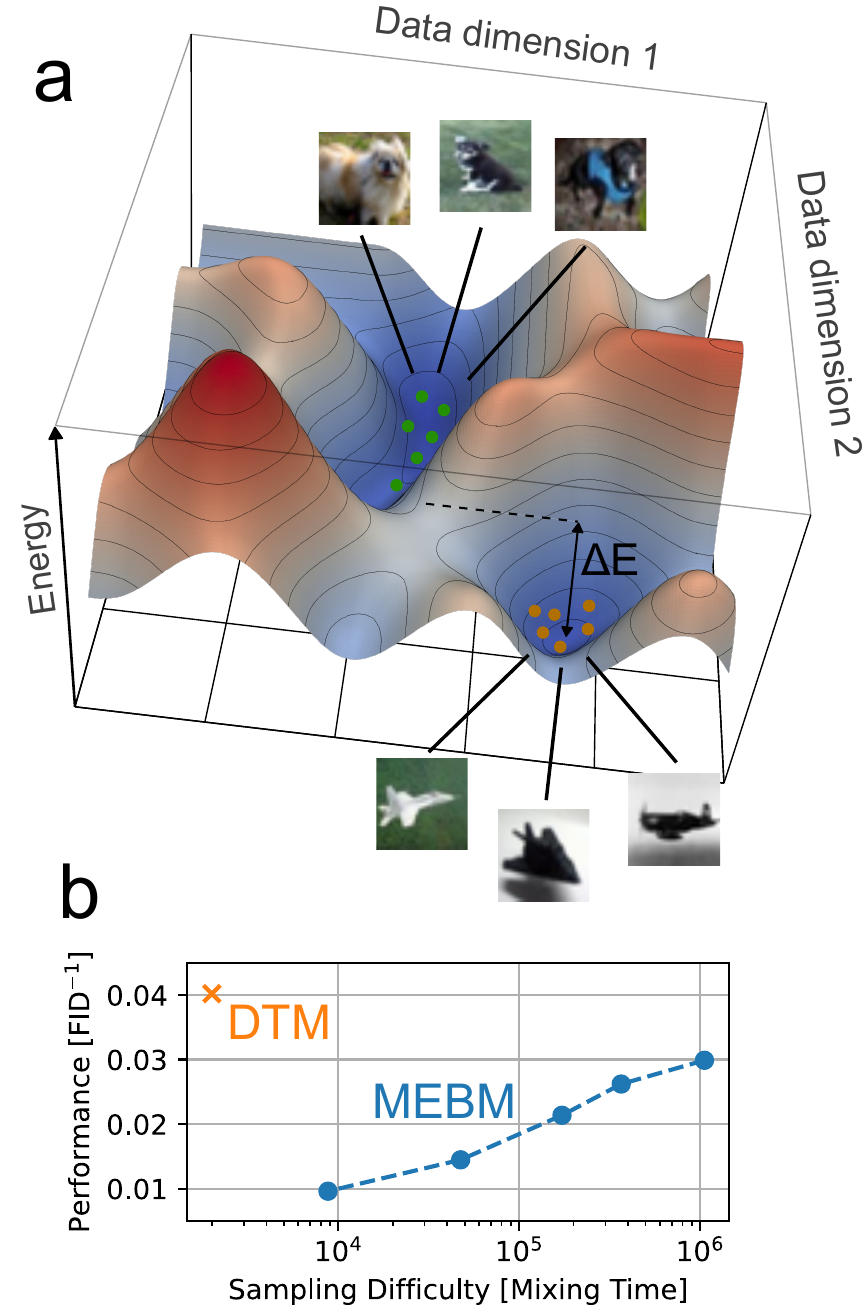}
\caption{\textbf{The mixing–expressivity tradeoff.} \textbf{(a)} A cartoon illustrating the mixing–expressivity tradeoff in EBMs. It shows a projection of an energy landscape fit to a simple dataset. The ``airplane'' mode is well separated from the ``dog'' mode, with very little data in between. Progressively better fits of the EBM to the data tend to feature larger energy barriers $\Delta E$ between the modes, making the EBM increasingly difficult to sample from. \textbf{(b)} An example of the effect of the mixing–expressivity tradeoff on model performance as measured using the Fashion-MNIST dataset. The blue curve in the plot shows the results of experiments on MEBMs with limited allowed mixing time. Performance and mixing time are strongly correlated. Mixing times were computed by fitting an exponential function to the large-lag behavior of the autocorrelation function; see Appendix~\ref{sec:mebm}. In contrast, a DTM (orange cross) has higher performance despite substantially lower sampling requirements.}
\label{fig:mixing_exp}
\end{figure}

Fitting an MEBM corresponds to assigning low energies to values of $x$ where data are abundant and high energies to values of $x$ that are far from data. Real-world data are often clustered into distinct modes~\cite{dempster1977maximum, bishop1994mixture}, meaning that an MEBM that fits data well will have a complex, rugged energy landscape with many deep valleys surrounded by tall mountains. This complexity is illustrated by the cartoon in Fig.~\ref{fig:mixing_exp}~(a).

Unlike the systems we propose, existing probabilistic computers based on MEBMs struggle with the multimodality of real-world data, which hinders their efficiency. Namely, the amount of energy the computer must expend to draw a sample from the MEBM's distribution can be tremendous if its energy landscape is very rough.

Specifically, sampling algorithms that operate in high dimensions (such as Gibbs sampling~\cite{pml2Book}) are locally informed iterative procedures, meaning that they sample a landscape by randomly making small movements in the space based on low-dimensional information. When using such a procedure to sample from Eq.~\eqref{eq:ebm}, the probability that the iteration will move up in energy to some state $X[k+1]$ is exponentially small in the energy increase compared to the current state $X[k]$, i.e.,
\begin{equation}
\mathbb{P}(X[k+1]= x' | X[k] = x) \propto e^{-\left( \mathcal{E}(x') - \mathcal{E}(x)\right)} .
\end{equation}
For large differences in energy, like those encountered when trying to move between two valleys separated by a significant barrier, this probability can be very close to zero. In standard Markov-chain analyses, such barriers lead to exponentially large expected transition times between modes, which is reflected in a rapidly growing mixing time.

The mixing–expressivity tradeoff (MET) summarizes this issue with existing probabilistic computer architectures, reflecting the fact that modeling performance and sampling hardness are coupled for MEBMs. Specifically, as the expressivity (modeling performance) of an MEBM increases, its \emph{mixing time} (the amount of computational effort needed to draw independent samples from the MEBM's distribution) becomes progressively longer, resulting in expensive inference and unstable training~\cite{carbone2023efficient, desjardins2010parallel}; see Appendix~\ref{app:autocorr} for a definition of mixing time.

The empirical effect of the MET on the efficiency of MEBM-based probabilistic computing systems is illustrated in Fig.~\ref{fig:mixing_exp}~(b). Mixing time increases very rapidly with performance, inflating the amount of energy required to sample from the model. The effect of this increased mixing time is reflected in Fig.~\ref{fig:perf}: despite the MEBM-based solution using the same EBMs and underlying hardware as the DTM-based solution, its energy consumption is several orders of magnitude larger due to the glacially slow mixing.

\section{Denoising thermodynamic models}

The MET makes it clear that MEBMs have a flaw that makes them challenging and energetically costly to scale. However, this flaw is avoidable, and many types of probabilistic machine learning models have been developed to solve the distribution modeling problem while circumventing the MET.

Denoising diffusion models were explicitly designed to sidestep the MET by gradually building complexity through a series of simple, easy-to-sample probabilistic transformations~\cite{diff_orig}. By doing so, they allowed for much more complex distributions to be expressed given a fixed compute budget and substantially expanded the capabilities of generative models~\cite{ho2022cascaded, rombach2022high, saharia2022photorealistic}.

DTMs merge EBMs with diffusion models, offering an alternative path for probabilistic computing that mitigates the MET. DTMs are a slight generalization of recent work from deep learning practitioners that has pushed the frontier of EBM performance~\cite{gao2021DRL, yu2022latent, xu2024energy, zhulearning}.

Instead of trying to use a single EBM to model the data, DTMs chain many EBMs to gradually build up to the complexity of the data distribution. This gradual buildup of complexity allows the landscape of each EBM in the chain to remain relatively simple (and easy to sample) without limiting the complexity of the distribution modeled by the chain as a whole; see Fig.~\ref{fig:mixing_exp}~(b).

Denoising models attempt to reverse a process that gradually transforms the data distribution $Q(x^0)$ into simple noise. This forward process is given by the Markov chain
\begin{equation}\label{eq:forward_process}
Q(x^0, \dots, x^T) = Q(x^0) \prod_{t=1}^T Q(x^t|x^{t-1}) .
\end{equation}
The forward process is typically chosen such that it has a unique stationary distribution $Q(x^T)$, which takes a simple form (e.g., Gaussian or uniform).

Reversal of the forward process is achieved by learning a set of distributions $P_{\theta}(x^{t-1}|x^t)$ that approximate the reversal of each conditional in Eq.~\eqref{eq:forward_process}. In doing so, we learn a map from simple noise to the data distribution, which can then be used to generate new data.

In traditional diffusion models, the forward process is made to be sufficiently fine-grained (using a large number of steps $T$) such that the conditional distribution of each step in the reverse process takes some simple form (such as Gaussian or categorical). This simple distribution is parameterized by a neural network, which is then trained to minimize the Kullback–Leibler (KL) divergence between the joint distributions $Q$ and $P_{\theta}$,
\begin{equation}\label{eq:denoising_loss_body}
\mathcal{L}_{DN}(\theta) = D\left( Q(x^0, \dots, x^T) \middle\| P_{\theta}(x^0, \dots, x^T)\right),
\end{equation}
where the joint distribution of the model is the product of the learned conditionals:
\begin{equation}
P_{\theta}(x^0, \dots, x^T) = Q(x^T) \prod_{t=1}^T P_{\theta}(x^{t-1}|x^t) .
\end{equation}
See Appendix~\ref{app:revproc} for more details.

EBM-based denoising models approach the problem from a different angle~\cite{gao2021DRL}. In many cases, it is straightforward to re-cast the forward process in an exponential form,
\begin{equation}\label{eq:forward_energy}
Q(x^t|x^{t-1}) \propto  e^{-\mathcal{E}^f_{t-1} \left( x^{t-1}, x^t\right)},
\end{equation}
where $\mathcal{E}^f_{t-1}$ is the energy function associated with the forward process step that adds noise to $x^{t-1}$. We then use an EBM with a particular energy function to model the conditional, i.e.,
\begin{equation}\label{eq:ebm_cond}
P_{\theta}(x^{t-1}|x^t) \propto e^{-\left( \mathcal{E}^f_{t-1} \left( x^{t-1}, x^t\right) + \mathcal{E}^{\theta}_{t-1}\left(x^{t-1}, \theta\right)\right)}.
\end{equation}
In this parameterization, the dependence on $x^t$ enters entirely through the forward energy $\mathcal{E}^f_{t-1}(x^{t-1}, x^t)$, which constrains $x^{t-1}$ to stay close to the noisy input $x^t$, while $\mathcal{E}^{\theta}_{t-1}$ shapes the local structure of the distribution over $x^{t-1}$.

Equation~\eqref{eq:ebm_cond} allows for a compromise between the number of steps in the approximation to the reverse process and the difficulty of sampling at each step. As the number of steps in the forward process is increased, the effect of each noising step becomes smaller, meaning that $\mathcal{E}^f_{t-1}$ more tightly binds $x^t$ to $x^{t-1}$. This binding can simplify the distribution given in Eq.~\eqref{eq:ebm_cond} by imposing an energy penalty that prevents it from being strongly multimodal; see Appendix~\ref{app:simple-energy-landscape} for further discussion.

\begin{figure}
\centering
\includegraphics[width=0.8\linewidth]{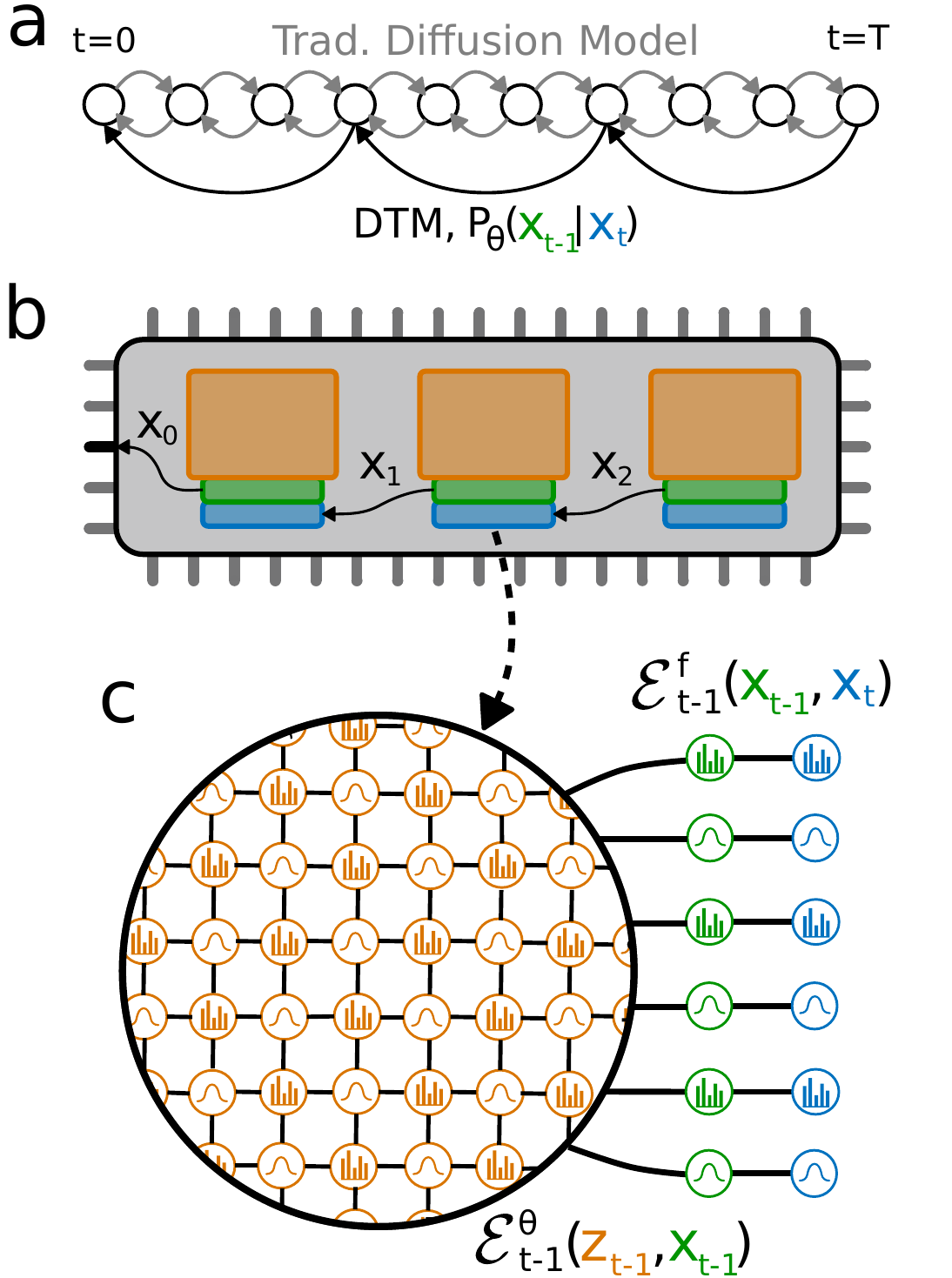}
\caption{\textbf{The denoising thermodynamic computer architecture.} \textbf{(a)} Traditional diffusion models have simple conditionals and must take small steps when approximating the reverse process. Since EBMs can express more complex distributions, DTMs can take potentially much larger steps. \textbf{(b)} A sketch of how a chip based on the DTCA chains hardware EBMs to approximate the reverse process. Each EBM is implemented by distinct circuitry, parts of which are dedicated to receiving the inputs and conditionally sampling the outputs and latents. \textbf{(c)} An abstract diagram of a hardware EBM. The state variables $x^{t}$ and $x^{t-1}$ map onto distinct physical degrees of freedom represented by the blue and green nodes, respectively. The coupling between these two sets of nodes implements the forward process energy function $\mathcal{E}_{t}^{f} \left( x^{t-1}, x^t\right)$. The set of orange nodes represents a set of latent variables $z^{t-1}$. The couplings between these nodes and to the $x^{t-1}$ nodes implements $\mathcal{E}_{t-1}^{\theta} \left( z^{t-1}, x^{t-1}\right)$.}
\label{fig:denoising}
\end{figure}

As illustrated in Fig.~\ref{fig:denoising}~(a), models of the form given in Eq.~\eqref{eq:ebm_cond} reshape simple noise into an approximation of the data distribution. Increasing $T$ while holding the EBM architecture constant simultaneously increases the expressive power of the chain and makes each step easier to sample from, entirely bypassing the MET.

To maximally leverage probabilistic hardware for EBM sampling, DTMs generalize Eq.~\eqref{eq:ebm_cond} by introducing latent variables $\{ z^t \}$:
\begin{equation}\label{eq:latent_cond}
P_{\theta}(x^{t-1}|x^t) \propto \sum_{z^{t-1}} e^{-\left( \mathcal{E}^f_{t-1} \left( x^{t-1}, x^t\right) + \mathcal{E}^{\theta}_{t-1}\left(x^{t-1}, z^{t-1}, \theta\right)\right)} .
\end{equation}
Introducing latent variables allows the size and complexity of the probabilistic model to be increased independently of the data dimension. In this generalized form, $\mathcal{E}^f_{t-1}$ still enforces proximity between $x^{t-1}$ and $x^t$, while $\mathcal{E}^{\theta}_{t-1}$ uses the latent variables $z^{t-1}$ to enrich the local conditional structure without directly depending on $x^t$.

A convenient property of DTMs is that if the approximation to the reverse-process conditional is exact ($P_{\theta}(x^{t-1}|x^t) \to Q(x^{t-1}|x^t)$), one also learns the marginal distribution at $t-1$,
\begin{equation}
Q(x^{t-1}) \propto \sum_{z^{t-1}} e^{-\mathcal{E}^{\theta}_{t-1}\left(x^{t-1}, z^{t-1}, \theta\right)} .
\end{equation}
See Appendix~\ref{app:learning-the-marginal} and Ref.~\cite{gao2021DRL} for further details and discussion of this property. Note that it relies on the normalizing constant associated with the distribution in Eq.~\eqref{eq:forward_energy} being independent of $x^{t-1}$.

\section{Denoising Thermodynamic Computers}

The Denoising Thermodynamic Computer Architecture (DTCA) tightly integrates DTMs into probabilistic hardware, allowing for the highly efficient implementation of EBM-aided diffusion models.

Practical implementations of the DTCA utilize natural-to-implement EBMs that exhibit sparse and local connectivity, as is typical in the literature~\cite{niazi2024training}. This constraint allows sampling of the EBM to be performed by massively parallel arrays of primitive circuitry that implement Gibbs sampling. Refer to Appendices \ref{app:hardwardacc} and \ref{sec:hw} for further theoretical discussion of the hardware architecture.

A key feature of the DTCA is that $\mathcal{E}^f_{t-1}$ can be implemented efficiently using our constrained EBMs. Specifically, for both continuous and discrete diffusion, $\mathcal{E}^f_{t-1}$ can be implemented using a single pairwise interaction between corresponding variables in $x^t$ and $x^{t-1}$; see Appendices~\ref{app:forward} and \ref{app:forward_ebm} for details. This structure can be reflected in how the chip is laid out to implement these interactions without violating locality constraints.

Critically, Eq.~\eqref{eq:latent_cond} places no constraints on the form of $\mathcal{E}_{t-1}^{\theta}$. Therefore, we are free to use EBMs that our hardware implements especially efficiently. At the lowest level, this corresponds to high-dimensional, regularly structured latent-variable EBMs. If more powerful models are desired, these hardware latent-variable EBMs can be arbitrarily scaled by combining them into software-defined graphical models.

The modular nature of DTMs enables various hardware implementations. For example, each EBM in the chain can be implemented using distinct physical circuitry on the same chip, as shown in Fig.~\ref{fig:denoising}~(b). Alternatively, the various EBMs may be split across several communicating chips or implemented by the same hardware, reprogrammed with distinct sets of weights at different times. For any given EBM in the chain, both the data variables $x^t$, $x^{t-1}$ and the latent variables $z^{t-1}$ are physically embodied in sampling circuits that are connected in a simple way that reflects the structure of Eq.~\eqref{eq:ebm_cond}. This variable structure is shown schematically in Fig.~\ref{fig:denoising}~(c).

To understand the performance of a future hardware device, we developed a GPU simulator of the DTCA and used it to train a DTM on the Fashion-MNIST dataset. We measure the performance of the DTM using FID and utilize a physical model to estimate the energy required to generate new images. These numbers can be compared to conventional algorithm/hardware pairings, such as a VAE running on a GPU; these results are shown in Fig.~\ref{fig:perf}.

The DTM that produced the results shown in Fig.~\ref{fig:perf} used Boltzmann machine EBMs. Boltzmann machines, also known as Ising models in physics, use binary random variables and are the simplest type of discrete-variable EBM.

Boltzmann machines are hardware efficient because the Gibbs sampling update rule required to sample from them is simple. Boltzmann machines implement energy functions of the form
\begin{equation}
\mathcal{E}(x) = -\beta \Big(\sum_{i \neq j} x_i J_{ij} x_j + \sum_{i=1} h_i x_i\Big) ,
\end{equation}
where each $x_i \in  \{ -1, 1 \}$. The Gibbs sampling update rule for sampling from the corresponding EBM is
\begin{equation}\label{eq:boltz_gibbs}
\mathbb{P}\left(X_i[k+1] = +1 \mid X[k] = x\right)
= \sigma\!\bigg(2\beta\Big(\sum_{j\neq i}J_{ij}\,x_j + h_i\Big)\bigg),
\end{equation}
which can be evaluated simply using an appropriately biased source of random bits.

\begin{figure*}[t]
\centering
\includegraphics[width=\linewidth]{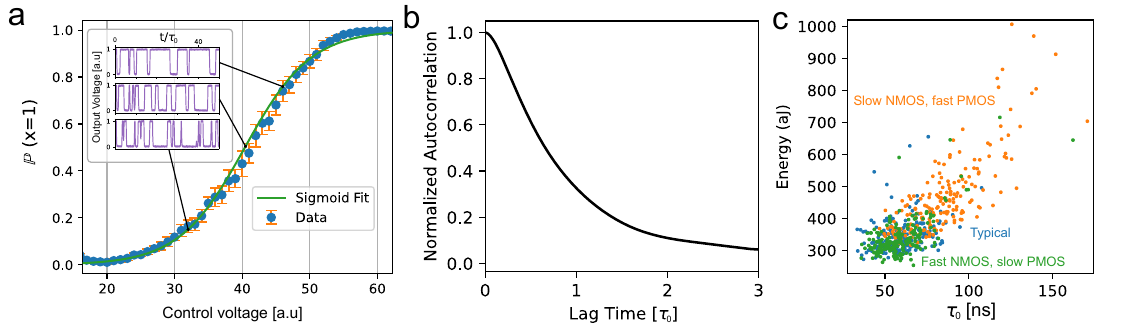}
\caption{\textbf{A programmable source of random bits. \textbf{(a)}} A laboratory measurement of the operating characteristic of our RNG. The probability of the output voltage signal being in the high state ($x = 1$) can be programmed by varying an input voltage. The relationship between $\mathbb{P}(x=1)$ and the input voltage is well-approximated by a sigmoid function. The inset shows the output voltage signal as a function of time for different input voltages. \textbf{(b)} The autocorrelation function of the RNG at the unbiased point ($\mathbb{P}(x=1) = 0.5$). The decay is approximately exponential with the rate $\tau_0 \approx 100 \text{ns}$. \textbf{(c)} Estimating the effect of manufacturing variation on RNG performance. Each point in the plot represents the results of a simulation of an RNG circuit with transistor parameters sampled according to a procedure defined by the manufacturer's PDK. Each color represents a different process corner, each for which $\sim 200$ realizations of the RNG were simulated. The ``typical'' corner represents a balanced case, whereas the other two are asymmetric corners where the two types of transistors (NMOS and PMOS) are skewed in opposite directions. The slow NMOS and fast PMOS case is worst performing for us due to an asymmetry in our design.}
\label{fig:pbit}
\end{figure*}

Implementing our proposed hardware architecture using Boltzmann machines is particularly simple. A device will consist of nodes connected into a regular grid. Each node represents a single Bernoulli random variable $x_i$, implemented as a transistor-based sampling circuit (later referred to as RNG). The bias of a sampling circuit (the probability that it produces 1 as opposed to $-1$) is constrained to be a sigmoidal function of an input voltage, allowing its update probability to be conditioned by the values of its neighbors as given in Eq.~\eqref{eq:boltz_gibbs}. This can be implemented using a simple circuit that adds currents such as a resistor network (see Appendix~\ref{sec:biasing}).

Specifically, the EBMs employed in this work were sparse, deep Boltzmann machines comprising $L \times L$ grids of binary variables, where $L = 70$ was used in most cases. Each variable was connected to several (in most cases, 12) of its neighbors following a simple pattern. At random, some of the variables were selected to represent the data $x_{t-1}$, and the rest were assigned to the latent variables $z_{t-1}$. Then, an extra node was connected to each data node to implement the coupling to $x_{t}$. See Appendix~\ref{sec:hw} for further details on the Boltzmann machine architecture.

Due to our chosen connectivity patterns, our Boltzmann machines are bipartite (two-colorable), meaning that nodes can be separated into two blocks, such that no two nodes within the same block are connected to each other. Since each color block can be sampled in parallel, a single iteration of Gibbs sampling corresponds to first sampling one color block conditioned on the other and then swapping their roles. Therefore, the total duration of one full Gibbs iteration (so updating all nodes) takes roughly $2\tau_0$ in wall-clock time, where $\tau_0$ is the decorrelation time of an individual RNG, that is the time it takes for a sampling circuit inside a single node of the Boltzmann machine to equilibrate. It should be noted that $\tau_0$ is set by the intrinsic stochastic dynamics of the subthreshold circuit and its operating point. It is not a freely tunable algorithmic parameter, but can only be modified indirectly through circuit redesign.

Starting from some random initialization, this block-sampling procedure can be repeated for $K$ iterations (where $K$ is longer than the mixing time of the sampler, typically $K \approx 1000$) to draw samples from Eq.~\eqref{eq:ebm_cond} for each step in the approximation to the reverse process. For a DTM running on a DTCA chip, the total time required to draw a single sample is therefore proportional to $T K \, \tau_0$, where $T$ is the number of denoising steps (i.e., EBMs that the DTM is comprised of). It is important to distinguish $\tau_0$ from $K_\text{mix}$. The former is an entirely hardware-related parameter, while the latter is the number of Gibbs iterations that the abstract Boltzmann machine (irrespective of hardware implementation) needs to approximately converge to its equilibrium distribution.

To enable a near-term, large-scale realization of the DTCA, we leveraged the shot-noise dynamics of subthreshold transistors~\cite{freitas2025} to build an RNG that is fast, energy-efficient, and small. Our all-transistor RNG is programmable and has the desired sigmoidal response to a control voltage, as shown by experimental measurements in Fig.~\ref{fig:pbit}~(a). The stochastic voltage signal output from the RNG has an approximately exponential autocorrelation function that decays in approximately $100$ ns, as illustrated in Fig.~\ref{fig:pbit}~(b). As shown in Ref.~\cite{freitas2025}, this time constant is much larger than the lower limit imposed by the correlation time of the noise in our transistors. The RNG could, therefore, be made much faster via an improved design. Appendix~\ref{sec:our_rng_details} provides further details about our RNG.

In practice, semiconductor manufacturing is never perfectly precise. Each fabrication step introduces small fluctuations in transistor characteristics, producing global shifts across wafers as well as random differences between neighboring devices on the same chip. Because these variations can change how fast or efficiently a circuit operates, it is important to verify that our approach remains reliable under realistic fabrication conditions.
A practical advantage of our all-transistor RNG is that detailed and proven foundry-provided models can be used to study the effect of manufacturing variations on our circuit design. In Fig.~\ref{fig:pbit}~(c), we use this process development kit (PDK) to study the speed and energy consumption of our RNG as a function of both systematic inter-wafer skews to the transistor parameters (process corners) and the expected variation within a single chip. We find that the RNG works reliably despite these non-idealities, meaning it can be readily scaled to the massive grids required by the DTCA.

The energy estimates given in Fig.~\ref{fig:perf} for the probabilistic computer were constructed using a physical model of an all-transistor Boltzmann machine Gibbs sampler. The dominant contributions to this model are captured by the formula
\begin{equation}
E = T K_{\text{mix}} L^2  E_{\text{cell}},
\end{equation}
\begin{equation}
E_{\text{cell}} = E_{\text{rng}} + E_{\text{bias}}+ E_{\text{clock}} + E_{\text{comm}},
\end{equation}
where $E_{\text{rng}}$ comes from the data in Fig.~\ref{fig:pbit}~(c). The term $E_{\text{bias}}$ is estimated using a physical model of a possible biasing circuit, and $E_{\text{clock}}$ and $E_{\text{comm}}$ are derived from physical reasoning about the costs of the clock and inter-cell communications, respectively. $K_{\text{mix}}$ is the number of sampling iterations required to satisfactorily mix the chain for inference, which is generally less than the number of iterations used during training. $K_{\text{mix}} = 250$ was used for the DTM (see Appendix~\ref{sec:energy_complete}), while the mixing time measured in Fig.~\ref{fig:mixing_exp} was used for the MEBM.

This model is approximate, but it captures the underlying physics of a real device and provides a reasonable order-of-magnitude estimate of the actual energy consumption. Generally, given the same transistor process we used for our RNG and some reasonable selections for other free parameters of the model, we can estimate $E_{\text{cell}} \approx 2 \; \text{fJ}$. See Appendix~\ref{app:en-analysis} for an exhaustive derivation of this model.

We use a simple model for the energy consumption of the GPU that underestimates the actual values. We compute the total number of floating-point operations (FLOPs) required to generate a sample from the trained model and divide that by the FLOP/joule specification given by the manufacturer. See Appendix~\ref{app:energy-gpu} for further discussion.

\begin{figure*}[t]
\centering
\includegraphics[width=\textwidth]{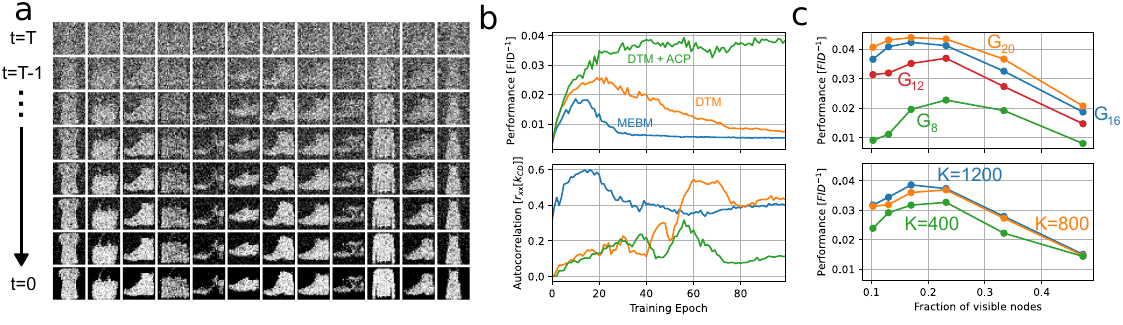}
\caption{\textbf{Detailed results on the Fashion-MNIST dataset.} \textbf{(a)} Images generated by a denoising model. Here, to achieve better-looking images, several binary variables were combined to represent a single grayscale pixel. The noisiness of the grayscale levels is an artifact of our embedding method; see Appendix~\ref{app:int-embed}. \textbf{(b)} An experiment showing how DTMs are more stable to train than MEBMs. Complementing DTMs with the ACP completely stabilizes training. For the DTMs, the maximum $r_{yy}[K]$ value over all the layers is shown. \textbf{(c)} The effect of scaling EBM complexity on DTM performance. The grid size $L$ was modified to change the number of latent variables compared to the (fixed) number of data variables. Generally, EBM layers with more connectivity and longer allowed mixing times can utilize more latent variables and, therefore, achieve higher performance.}
\label{fig:fmnist_main}
\end{figure*}

\section{Training DTMs}

The EBMs used in the experiments presented in Fig.~\ref{fig:perf} were trained by applying the standard Monte Carlo estimator for the gradients of EBMs~\cite{song2021train} to Eq.~\eqref{eq:denoising_loss_body}, which yields
\begin{align}\label{eq:mc_grad}
\begin{split}
    \nabla_{\theta} \mathcal{L}_{DN}(\theta) \!=\! \sum_{t=1}^T \mathbb{E}_{Q(x_{t-1}, x_t)}\bigg[&\mathbb{E}_{P_{\theta}(z_{t-1}|x_{t-1}, x_t)} \left[ \nabla_{\theta} \mathcal{E}^m_{t-1}\right] \\ - & \mathbb{E}_{P_{\theta}(x_{t-1}, z_{t-1}|x_t)} \left[ \nabla_{\theta} \mathcal{E}^m_{t-1}\right]  \bigg] .
\end{split}
\end{align}
Notably, each term in the sum over $t$ can be computed independently. To estimate either term in Eq.~\eqref{eq:mc_grad}, first, sample tuples $(x_{t-1}, x_t)$ from the forward process $Q(x_{t-1}, x_t)$. Then, for each of these tuples, clamp the reverse process EBM to the sampled values appropriately and use a time average over $K$ iterations of Gibbs sampling to estimate the inner expectation value. Averaging the result over the tuples yields the desired gradient estimate.

It should be noted that the DTCA allows our EBMs to have finite and short mixing times, which enables sufficient sampling iterations to be used to achieve nearly unbiased estimates of the gradient. Unbiased gradient estimates are not possible for MEBMs in most cases due to their long mixing times~\cite{carreira2005contrastive}.

A well-trained denoising model generates new examples that resemble the training data by incrementally pulling them out of noise; the outputs of an 8-step denoising model trained on the Fashion-MNIST dataset are shown in Fig.~\ref{fig:fmnist_main}~(a). At the final time $T$, the images are random bits. Structure begins to emerge as the chain progresses, ultimately resulting in clean images at time $ t=0$.

DTMs alleviate the training instability that is fundamental to MEBMs. The parameters of MEBMs are usually initialized using a strategy that results in an easy-to-sample-from energy landscape~\cite{hinton2012practical}. For this reason, in the early stages of training, sampling from Eq.~\eqref{eq:ebm} is possible, and the gradient estimates produced using Eq.~\eqref{eq:mc_grad} are unbiased. However, as these gradients are followed, the MEBM is reshaped according to the data distribution and begins to become complex and multimodal. This induced multimodality greatly increases the sampling complexity of the distribution, causing samples to deviate from equilibrium. Gradients computed using non-equilibrium samples do not necessarily point in a meaningful direction, which can halt or, in some cases, even reverse the training process.

This instability in MEBMs leads to unpredictable training dynamics that can be sensitive to implementation details. An example of the training dynamics for several different types of models is shown in Fig.~\ref{fig:fmnist_main}~(b). The top plot displays the quality of images generated during training, while the bottom plot shows a measure of the sampler's mixing. Image quality is measured using the FID metric, and mixing quality is measured using normalized autocorrelation $r_{yy}$ (see Appendix~\ref{app:autocorr}). The lower plot in Fig.~\ref{fig:fmnist_main}~(b) shows the autocorrelation at a delay equal to the total number of sampling iterations used to estimate the gradients during training. Generally, if $r_{yy}$ is close to 1, gradients were estimated using far-from-equilibrium samples and were likely of low quality. If it is close to zero, the samples should be close to equilibrium and produce high-quality gradient estimates. See Appendix~\ref{app:autocorr} for further discussion.

The destabilizing effect of non-equilibrium sampling is apparent from the blue curves in Fig.~\ref{fig:fmnist_main}~(b). At the beginning of training, both quality and $r_{yy}$ increase, indicating that the multimodality of the data is being imprinted on the model. As training progresses and the energy landscape becomes more rugged, $r_{yy}$ becomes so large that the quality of the gradient starts to decline, resulting in a plateau and, ultimately, a degradation of the model's quality.

Denoising alone significantly stabilizes training. Because the transformation carried out by each layer is simpler, the distribution that the model must learn is less complex and, therefore, easier to sample from. The orange curve in Fig.~\ref{fig:fmnist_main}~(b) shows the training dynamics for a typical denoising model. The autocorrelation and performance remain good for much longer than the MEBM.

As training progresses, the DTM eventually becomes unstable, which can be attributed to the development of a complex energy landscape among the latent variables. To combat this, we modify the training procedure to penalize models that mix poorly. We add a term to the loss function that nudges the optimization towards a distribution that is easy to sample from, i.e.,
\begin{equation}
\mathcal{L}^{TC}_t = \mathbb{E}_{Q(x^t)} \left[D\left(\prod_{i=1}^M P_{\theta}(s^{t-1}_i|x^t)  \middle\| P_{\theta}(s^{t-1}|x^t)\right) \right] ,
\end{equation}
where $s^{t-1} = (x^{t-1}, z^{t-1})$ and $x^{t-1}_i$ indicates the $i^{\text{th}}$ of the $M$ variables in $x^{t-1}$. This term penalizes the distance between the learned conditional distribution and a factorized distribution with identical marginals and is a form of total correlation penalty~\cite{chen2018isolating}. See Appendices~\ref{app:acp} and~\ref{app:sampling_time_and_acp} for further discussion of this penalty and its relationship to mixing time and sampling effort.

The total loss function is the sum of Eq.~\eqref{eq:denoising_loss_body} and this total correlation penalty:
\begin{equation}
\mathcal{L} = \mathcal{L}_{DN} + \sum_{t=1}^T \lambda_t \mathcal{L}^{TC}_t.
\end{equation}
The parameters $\lambda_t$ control the relative strength of the total correlation penalty for each step in the reverse process.

We use an Adaptive Correlation Penalty (ACP) to set the $\lambda_t$ as large as necessary to keep sampling tractable for each layer. During training, we periodically measure the autocorrelations of each learned conditional at a delay equal to the number of sampling iterations used during gradient estimation. If the autocorrelation for the $j^{\text{th}}$ layer is close to zero, $\lambda_j$ is decreased, and vice versa.

Our closed-loop control of the correlation penalty strengths is crucial, allowing us to maximize the expressivity of the EBMs while maintaining stable training. The green curves in Fig.~\ref{fig:fmnist_main}~(b) show an example of training dynamics under this closed-loop control policy. Model quality increases monotonically, and the autocorrelation stays small throughout training. This closed-loop control of the correlation penalty was employed during the training of most models used to produce the results in this article, including those shown in Fig.~\ref{fig:perf}. Further analysis of the ACP and its effect on mixing is given in Appendices~\ref{app:acp} and~\ref{app:sampling_time_and_acp}.

Generally, the performance of DTMs improves as their size increases. As shown in Fig.~\ref{fig:perf}, increasing the depth of the DTM from 2 to 8 substantially improves the quality of generated images. As shown in Fig.~\ref{fig:fmnist_main}~(c), increasing the width, degree, and allowed mixing time of the EBMs in the chain also generally improves performance.

However, some subtleties prevent this specific EBM topology from being scaled indefinitely. The top plot in Fig.~\ref{fig:fmnist_main}~(c) shows that scaling the number of latent variables (with fixed allowed mixing time) only improves performance if the connectivity of the graph is also scaled; otherwise, performance can decrease. This dependence makes sense, as increasing the number of latent variables in this way increases the depth of the Boltzmann machine, which is known to make sampling more difficult. Beyond a certain point, increasing the model's ability to express complex energy landscapes may render it unable to learn, given the allowed mixing time of $K \approx 1000$. This same effect is shown in the bottom plot of Fig.~\ref{fig:fmnist_main}~(c), which demonstrates that larger values of $K$ are required to support wider models while holding connectivity constant. Further study of how performance scales with sampling time $K$ and the number of denoising steps $T$ is provided in Appendix~\ref{app:sampling_time_and_acp}.

In general, it would be naive to expect that a hardware-efficient EBM topology can be scaled in isolation to model arbitrarily complex datasets. For example, there is no good reason why a connectivity pattern that is convenient from a wire-routing perspective would also be well suited to represent the correlation structure of a complex real-world dataset.

\section{Hybrid Thermodynamic-Deterministic Machine Learning}
\label{sec:htdml}

The core doctrine of modern machine learning is the relentless scaling of models as a means of solving ever-harder problems. Models that utilize probabilistic computers may be similarly scaled to enhance their capabilities beyond the relatively simple dataset considered in this work so far.

However, we hypothesize that the correct way to scale probabilistic machine learning hardware systems is not in isolation but rather as a component in a larger \emph{hybrid thermodynamic-deterministic machine learning} (HTDML) system. Such a hybrid system integrates probabilistic hardware with more traditional machine learning accelerators. This hybrid approach is particularly important for moving beyond binarized grayscale datasets such as Fashion-MNIST toward richer, higher-dimensional data.

A hybrid approach is sensible because there is no \emph{a priori} reason to believe that a probabilistic computer should handle every part of a machine learning problem, and sometimes a deterministic processor is likely a better tool for the job.

The goal of HTDML is to design practical machine learning systems that minimize the energy used to achieve desired modeling fidelity on a particular task. This efficiency will be achieved through a cross-disciplinary effort that eschews the software/hardware abstraction barrier to design computers that respect physical constraints and leverage each type of hardware where it is most effective.

For example, more rigorous methods of embedding data into hardware EBMs will need to be developed to go beyond the relatively simple datasets considered here. Indeed, binarization is not viable in general, and embedding into richer types of variables (such as categorical) at the probabilistic hardware level is not particularly efficient or principled. Instead, one may try a hybrid (HTDML) approach where a small neural network (the deterministic part) learns to embed the data into a binary latent space and a DTM (the thermodynamic part) can then be trained inside this binary space. A similar approach has been very beneficial in the context of diffusion models \cite{yu2022latent}, to the point that now many state-of-the-art diffusion models operate in the latent space of a pre-trained VAE.

\begin{figure}[t]
\centering
\includegraphics[width=\linewidth]{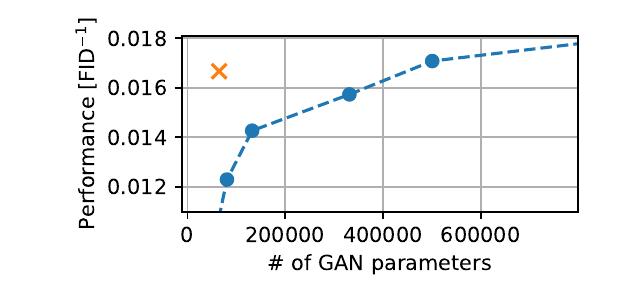}
\caption{\textbf{Embedding data into a DTM using a neural network.} Here, we show the results of using a simple embedding model in combination with a DTM. The DTM is trained to generate CIFAR-10 images and achieves performance parity with a traditional GAN using a $\sim 10\times$ smaller deterministic neural network in terms of parameter count.}
\label{fig:embedding}
\end{figure}

We show the results of a quick initial attempt at such hybrid models in Fig.~\ref{fig:embedding}. Here, we train a small neural network to embed the CIFAR-10 dataset~\cite{krizhevsky2009learning} into a binary DTM. The embedding network was trained using an autoencoder loss to binarize the data, which was then used to train a DTM. The decoder of the embedding network was then trained further using a GAN objective to increase the quality of the generated images. This training procedure is described in further detail in Appendix~\ref{app:determ-embed}. Example CIFAR-10 images generated by our model can be found in Appendix~\ref{app:cifar_imgs}.

Despite the overhead of the embedding neural network, this primitive hybrid model is efficient. As shown in the figure, to achieve equivalent performance with a purely NN-based GAN, the deterministic part of that GAN would need to have roughly ten times more parameters than the deterministic part of our hybrid model. Informally, one might say that therefore the DTM holds about 90\% of the expressivity of our model while the neural network accounts for only about 10\% of the parameters.

We should again stress that this is just a naive first attempt at a hybrid model, and we believe that further research into hybrid thermodynamic–deterministic models can lead to even more significant efficiency gains and scale this approach to larger tasks. For example, one major flaw of the embedding procedure employed in Fig.~\ref{fig:embedding} is that the autoencoder and DTM are not jointly trained, which means that the embedding learned by the autoencoder may not be well-suited to the way information can flow in the DTM, given its limited connectivity. Finding a good way to jointly train them seems like a promising future research direction.

\section{Conclusion}

In this paper, we have given an early view of how one might design generative machine learning algorithms and hardware jointly for thermodynamic probabilistic computers. At the modeling level, we introduced Denoising Thermodynamic Models (DTMs), which repurpose hardware EBMs as denoising steps rather than monolithic models of the data distribution and thereby avoid the mixing–expressivity tradeoff that limits monolithic EBMs. At the hardware level, we proposed the Denoising Thermodynamic Computer Architecture (DTCA), which implements DTMs using sparse, locally connected Boltzmann machines driven by an all-transistor RNG with a tunable bias and short autocorrelation time. By combining these ingredients with an adaptive correlation-penalty training procedure, we estimate that a DTCA device could match the performance of GPU-based generative models on binarized Fashion-MNIST while using roughly four orders of magnitude less energy per generated sample.

It is important to note that the experiments presented in this paper were done using simulations of thermodynamic hardware, running on a classical computer, which substantially limited the maximum size of the thermodynamic model we could represent. Specifically, the largest DTM shown in Fig.~\ref{fig:perf} uses only around \numprint{50000} nodes in total, which is much smaller than what we expect the capacity of early DTCA chips will be. In fact, based on the size of our RNG, it can be estimated that $\sim 10^6$ sampling cells could be fit into a $6 \times 6 \; \text{µm}$ chip (see Appendix~\ref{sec:our_rng_details}). Furthermore, measurements of our RNG suggest that early DTCA chips will be about two to three orders of magnitude faster than the simulations used in this article. As shown in \cref{fig:steps_cd_fid_phase_diagram}, using more Gibbs steps during training and inference as well as using more denoising steps both increase model performance (albeit with diminishing returns). Furthermore, as shown in \cref{fig:fmnist_main} (c, bottom), as the model gets larger, increasing the number of Gibbs steps has a larger impact on performance. Realizing large-scale probabilistic computers using advanced transistor processes~\cite{sekigawa1984calculated, wu20223nm, tsmc5} will therefore both increase the capabilities of the DTM algorithm presented here and substantially speed up future research into other thermodynamic algorithms. The present work should thus be viewed as establishing a qualitative scaling picture and design template rather than as an optimized end-point.

Given the gap between the size of our models and the capabilities of a potential hardware device, a natural question to study is how these probabilistic models can be scaled beyond the obvious approaches considered here. This scaling likely corresponds to developing architectures that fuse multiple EBMs to implement each step in the reverse process. One possible approach is to construct software-defined graphical models of EBMs that enable non-local information routing, which could alleviate some of the issues associated with a fixed and local interaction structure. Such research would go hand-in-hand with HTDML, since classical neural networks could be used to learn an embedding that hierarchically distributes the generative task between several EBMs in a way similar to nVAE \cite{vahdat2021nvaedeephierarchicalvariational} or StyleGAN \cite{karras2019stylebasedgeneratorarchitecturegenerative}. Our preliminary CIFAR-10 experiments already show that pairing a relatively small deterministic front-end with a DTM back-end can offload most of the generative burden onto the thermodynamic component.

To address the current unavailability of probabilistic hardware, we have open-sourced a software library~\cite{thrmlgithub} that enables XLA-accelerated~\cite{xla} simulation of hardware EBMs. This library is written in JAX~\cite{jax2018github} and automates the complex slicing operations that enable hardware EBM sampling. We also provide additional code that wraps this library to implement the specific experiments presented in this article~\cite{osscode}. By making these tools available, we aim to lower the barrier for others to explore alternative thermodynamic algorithms, hardware topologies, and hybrid architectures, and to stress-test and refine the DTCA design in a broader range of applications.

In short, DTMs and the DTCA provide a concrete path toward generative models and hardware that are co-designed for thermodynamic probabilistic computation, rather than retrofitted to existing digital accelerators.

\newpage

\section{Declarations}

\subsection{Data Availability and Code Availabilty}
The code which reproduces our results and data is available at \href{https://github.com/pschilliOrange/dtm-replication}{github.com/pschilliOrange/dtm-replication}.

\subsection{Acknowledgements}
This research was funded by Extropic Corp.

\subsection{Author Contributions}
AJ and TM contributed equally, with TM doing most of the writing and energy estimation work, whereas AJ contributed most to the algorithms for training and evaluating the models presented. OL evaluated the performance of competitive models that we were comparing against. AG helped design some of the mentioned integrated circuits. PS helped with some experiments, and formatted the research code (which was published on GitHub) to be easy to understand and use. IC advised TM at the outset of this project. GV provided suggestions at the idea stage.

\subsection{Competing Interests}
All authors are part of Extropic corp., which produces and aims to sell products using some of the technologies described in the paper.

\putbib[settings,main]   
\end{bibunit}

\clearpage

\begin{bibunit}
\nocite{apsrev42Control}

\appendix

\onecolumngrid

\makeatletter
\renewcommand\p@subsection{\thesection.}
\makeatother

\section{List of Acronyms}
\label{app:acronyms}

\begin{table}[h]
\centering
\begin{tabular}{ll}
\hline
Acronym & Meaning \\
\hline
AI      & Artificial Intelligence \\
ACP     & Adaptive Correlation Penalty \\
DTM     & Denoising Thermodynamic Model \\
DTCA    & Denoising Thermodynamic Computer Architecture \\
EBM     & Energy-Based Model \\
MEBM    & Monolithic Energy-Based Model \\
FID     & Fréchet Inception Distance \\
FLOP    & Floating-Point Operation \\
GAN     & Generative Adversarial Network \\
GPU     & Graphics Processing Unit \\
HTDML   & Hybrid Thermodynamic–Deterministic Machine Learning \\
LLM     & Large Language Model \\
NN      & Neural Network \\
PDK     & Process Development Kit \\
RNG     & Random-Number Generator \\
VAE     & Variational Autoencoder \\
DDPM    & Denoising Diffusion Probabilistic Model \\
\hline
\end{tabular}
\caption{Acronyms used throughout this work.}
\end{table}

\section{Denoising Diffusion Models }

Denoising diffusion models try to learn to time-reverse a random process that converts data into simple noise. Here, we will review some details on how these models work to support the analysis in the main text.

\subsection{Forward Processes}\label{app:forward}

The forward process is a random process that is used to convert the data distribution into noise. This conversion into noise is achieved through a stochastic differential equation in the continuous-variable case and a Markov jump process in the discrete case.

\subsubsection{Continuous Variables}

In the continuous case, the typical choice of forward process is the It\^{o} diffusion,
\begin{equation}\label{eq:cont_diff}
dX(t) = -X(t) dt + \sqrt{2} \sigma d W
\end{equation}
where $X(t)$ is a length $N$ vector representing the state variable at time $t$, $\sigma$ is a constant, and $d W$ is a length $N$ vector of independent Wiener processes.

The transition kernel for a random process defines how the probability distribution evolves in time,
\begin{equation}\label{eq:trans_ker}
Q_{t|0}(x'|x) = \mathbb{P}(X(t) = x' | X(0) = x)
\end{equation}
For the case of Eq.~\eqref{eq:cont_diff} the transition kernel is,
\begin{equation}
Q_{t+s|s}(x'|x) \propto e^{-\frac{1}{2} (x' - \mu)^T \Sigma^{-1} (x' - \mu)}
\end{equation}
\begin{equation}
\mu = e^{-t} x
\end{equation}
\begin{equation}
\Sigma = \sigma^2 I \left( 1 - e^{-2t}\right)
\end{equation}
this solution can be verified by direct substitution into the corresponding Fokker-Planck equation. In the limit of infinite time, $\mu \to 0$ and $\Sigma \to \sigma^2 I$. Therefore, the stationary distribution of this process is zero-mean Gaussian noise with a standard deviation of $\sigma$.

\subsubsection{Discrete Variables}
\label{subsub:forward_proc_discrete}

The stochastic dynamics of some discrete variable $X$ may be described by the Markov jump process,

\begin{equation}\label{eq:mjp}
\frac{d Q_t}{d t} = \mathcal{L} Q_t
\end{equation}
where $\mathcal{L}$ is the generator of the dynamics, which is an $M \times M$ matrix that stores the transition rates between the various states. $Q_t$ is a length $M$ vector that assigns a probability to each possible state $X$ may take at time $t$. 

The transition rate from the $i^{\text{th}}$ state to the $j^{\text{th}}$ state is given by the matrix element $\mathcal{L}\left[ j, i\right]$, which here takes the particular form,
\begin{equation}\label{eq:L}
\mathcal{L}\left[ j, i\right] = \gamma \left(-(M - 1) \delta_{j, i} +(1 - \delta_{j, i})\right)
\end{equation}
where $\delta$ is used to indicate the Kronecker delta function. Eq.~\eqref{eq:L} describes a random process where the probability per unit time to jump between any two states is $\gamma$. 

Since Eq.~\eqref{eq:mjp} is linear, the dynamics of $Q_t$ can be understood entirely via the eigenvalues and eigenvectors of $\mathcal{L}$,
\begin{equation}
\mathcal{L} v_k = \lambda_k v_k
\end{equation}
Note that by symmetry of $\mathcal{L}$, we do not distinguish between right and left eigenvectors.

One eigenvector-eigenvalue pair $\left(v_0, \lambda_0=0\right)$ corresponds to the unique stationary state of $\mathcal{L}$, with all entries of $v_0$ being equal to some constant (if normalized, then $v_0[j] = \frac{1}{M}$ for all $j$).
The long-time dynamics of this MJP transform any initial distribution to a uniform distribution over all states.

The remaining eigenvectors are decaying modes associated with negative eigenvalues. These additional $M-1$ eigenvectors take the form,
\begin{equation}\label{eq:decay_vec}
v_j[i] = -\delta_{i, 0} + \delta_{i, j}
\end{equation}
\begin{equation}\label{eq:decay_val}
\lambda_j = -\gamma M
\end{equation}
where Eq.~\eqref{eq:decay_vec} and Eq.~\eqref{eq:decay_val} are valid for $j \in [1, M-1]$. Therefore, all solutions to this MJP decay exponentially to the uniform distribution with rate $\gamma M$.

The time-evolution of $Q$ is given by the matrix exponential,
\begin{equation}
Q_t = e^{\mathcal{L} t} Q_0.
\end{equation}

This matrix exponential is evaluated by diagonalizing $\mathcal{L}$,
\begin{equation}
e^{\mathcal{L} t} = P e^{D t} P^{-1}
\end{equation}
where the columns of $P$ are the $M$ eigenvectors $v_k$ and $D$ is a diagonal matrix of the eigenvalues $\lambda_k$.

Using the solution for the eigenvalues and eigenvectors found above, we can solve for the matrix elements of $e^{\mathcal{L} t}$,
\begin{equation}
e^{\mathcal{L} t} \left[ j, i\right] = \delta_{i, j} \left( \frac{1 + (M-1) e^{-\gamma M t}}{M}\right) + (1-\delta_{i, j}) \left(\frac{1 - e^{-\gamma M t}}{M}\right)
\end{equation}

Using this solution, we can deduce an exponential form for the matrix elements of $e^{\mathcal{L} t}$,
\begin{equation}\label{eq:gen_exp}
e^{\mathcal{L} t} \left[ j, i\right] = \frac{1}{Z(t)} e^{\Gamma(t) \delta_{i, j}}
\end{equation}
\begin{equation}\label{eq:gen_gamma}
\Gamma(t) = \ln{\left(\frac{1 + (M-1) e^{-\gamma t}}{1 - e^{-\gamma t}}\right)}
\end{equation}
\begin{equation}\label{eq:gen_part}
Z(t) = \frac{M}{1 - e^{-\gamma t}}
\end{equation}

Now consider a process in which each element of the vector of $N$ discrete variables $X$ undergoes the dynamics described by Eq.~\eqref{eq:mjp} independently. In that case, the differential equation describing the dynamics of the joint distribution $Q_t$ is,
\begin{equation}\label{eq:disc_diff_eq}
\frac{d Q_t}{d t} = \sum_{k=1}^N \left(I_1 \otimes \dots \otimes \mathcal{L}_k \otimes \dots I_N \right) Q_t
\end{equation}
where $I_j$ indicates the identity operator and $\mathcal{L}_j$ the operator from Eq.~\eqref{eq:L} acting on the subspace of the $j^{\text{th}}$ discrete variable. 

The Kronecker product of the matrix exponentials gives the time-evolution of the joint distribution,
\begin{equation}
e^{\mathcal{L} t} = \bigotimes_{k=1}^N e^{\mathcal{L}_k t}
\end{equation}
with the matrix elements,
\begin{equation}
e^{\mathcal{L} t}[j, i] = \prod_{k=1}^N e^{\mathcal{L}_k t}[j_k, i_k]
\end{equation}
where $j$ and $i$ are now vectors with $N$ elements, indexed as $i_k$ or $j_k$ respectively.

Using Eqs.~\eqref{eq:gen_exp} - \eqref{eq:gen_part}, we can find an exponential form for the joint process transition kernel (as defined in Eq.~\eqref{eq:trans_ker}),
\begin{equation}\label{eq:disc_rev_trans_ker}
Q_{t|0}(x'|x) = \frac{1}{Z(t)} \exp{\left(\sum_{k=1}^N \Gamma_k(t) \delta_{x'[k], x[k]}\right)}
\end{equation}
\begin{equation}
Z(t) = \prod_{k=1}^N Z_k(t)
\end{equation}
$\Gamma_k(t)$ and $Z_k(t)$ are as given in Eqs.~\eqref{eq:gen_gamma} and \eqref{eq:gen_part}, with each dimension potentially having it's own transition rate $\gamma_k$ and number of categories $M_k$.

\subsection{Reverse Processes}\label{app:revproc}

In general, a random process for some variable $X$ can be reversed using Bayes' rule,
\begin{equation}\label{eq:rev}
Q_{t|t+\Delta t}(x'|x) = \frac{Q_{t+\Delta t|t}(x|x') Q_t(x')}{Q_{t+\Delta t}(x)}
\end{equation}
where the conditionals are as defined in Eq.~\eqref{eq:trans_ker}, and the marginals are,
\begin{equation}
Q_t(x) = \mathbb{P}\left( X(t) = x\right)
\end{equation}

A differential equation that describes the reverse process can be found by analyzing Eq.~\eqref{eq:rev} in the infinitesimal time limit. Specifically, defining a reversed time $t = T - s$ given some arbitrary endpoint $T$ and expanding Eq.~\eqref{eq:rev} in $\Delta s$,
\begin{equation}\label{eq:rev_ker}
Q_{T-s|T-(s - \Delta s)}(x'|x) \approx \delta_{x, x'} + \Delta s \: \mathcal{L}_{\text{rev}}(x', x) 
\end{equation}
where $\mathcal{L}_{\text{rev}}$ is the generator of the reverse process,
\begin{equation}\label{eq:rev_gen}
\mathcal{L}_{\text{rev}}(x', x) =  \frac{Q_{T-s}(x')}{Q_{T-s}(x)} \lim_{\Delta s \to 0} \left[\frac{d}{d \Delta s} Q_{T - (s-\Delta s)|T-s}(x|x') \right] + \delta_{x, x'} \left( \frac{1}{Q_{T-s}(x) }\frac{d Q_{T-s}(x)}{ds}\right)
\end{equation}
here, $\delta_{x, x'}$ is used to indicate the Dirac delta function in the continuous case and the Kronecker delta in the discrete case.

If the dynamics of Q are linear and generated by $\mathcal{L}$ like  Eq.~\eqref{eq:mjp}, we can simplify,
\begin{equation}
\lim_{\Delta s \to 0} \left[\frac{d}{d \Delta s} Q_{T - (s-\Delta s)|T-s}(x|x') \right] = \mathcal{L}(x, x')
\end{equation}
In this case, we can re-write Eq.~\eqref{eq:rev_gen} in the operator form,
\begin{equation}\label{eq:rev_op}
\mathcal{L}_{\text{rev}} = Q \mathcal{L}^{\dagger} Q^{-1} + Q^{-1} \frac{d Q}{d s}
\end{equation}
where $\mathcal{L}^{\dagger}$ is the adjoint operator to $\mathcal{L}$. For continuous variables, the adjoint operator is defined as,
\begin{equation}\label{eq:def_adjoint}
\int \psi_2 \mathcal{L} \psi_1 dx = \int \psi_1 \mathcal{L}^{\dagger} \psi_2 dx
\end{equation}
for any test functions $\psi_1$ and $\psi_2$. For discrete variables, $\mathcal{L}^{\dagger} = \mathcal{L}^T$.

\subsubsection{Continuous variables}

In the case that the forward process is an It\^{o} diffusion, $\mathcal{L}$ is the generator for the corresponding Fokker-Planck equation,
\begin{equation}
\mathcal{L} = -\sum_i \frac{\partial}{\partial x_i} f_{i}(x, t) + \frac{1}{2} \sum_{i, j} \frac{\partial}{\partial x_i} \frac{\partial}{\partial x_j} D_{ij}(t)
\end{equation}
where $D$ is a symmetric matrix, $D_{ij} = D_{ji}$ that does not depend on $x$.

Using Eq.~\eqref{eq:def_adjoint} and integration by parts, it can be shown that the adjoint operator is,
\begin{equation}\label{eq:ito_adj}
\mathcal{L}^{\dagger} = \sum_i f_i \frac{\partial}{\partial x_i} + \frac{1}{2} \sum_{i, j} D_{ij} \frac{\partial}{\partial x_i} \frac{\partial}{\partial x_i}
\end{equation}
By directly substituting Eq.~\eqref{eq:ito_adj} into Eq.~\eqref{eq:rev_op} and simplifying, $\mathcal{L}_{\text{rev}}$ can be reduced to,
\begin{equation}
\mathcal{L}_{\text{rev}} = \sum_i \frac{\partial}{\partial x_i} g_i + \frac{1}{2} \sum_{i,j} \frac{\partial}{\partial x_i} \frac{\partial}{\partial x_j}  D_{ij}
\end{equation}
with the drift vector $g$,
\begin{equation}\label{eq:rev_fp}
g_i(x, t) = f_i(x, t) - \frac{1}{Q_t(x)} \sum_j \frac{\partial}{\partial x_j} \left[ D_{ij}(x, t) Q_t(x)\right]
\end{equation}

If $\Delta t$ is chosen to be sufficiently small,  Eq.~\eqref{eq:rev_fp} can be linearized and the transition kernel is Gaussian,
\begin{equation}
Q_{t|t+\Delta t}(x'|x) \propto \exp{\left(-\frac{1}{2} (x - \mu)^T \Sigma^{-1} (x - \mu) \right)}
\end{equation}
\begin{equation}
\mu = x + \Delta t \: g_i(x, t)
\end{equation}
\begin{equation}
\Sigma = \Delta t \: D(t)
\end{equation}
Therefore, one can build a continuous diffusion model with arbitrary approximation power by working in the small $\Delta t$ limit and approximating the reverse process using a Gaussian distribution with a neural network defining the mean vector~\cite{sohl2015deep, ho2020denoising}.

\subsubsection{Discrete variables}

In a discrete diffusion model, $\mathcal{L}$ is given by Eq.~\eqref{eq:disc_diff_eq}. This tensor product form for $\mathcal{L}$ guarantees that $\mathcal{L}(x', x) = 0$ for any vectors $x'$ and $x$ that have a Hamming distance greater than one (which means they have at least $N-1$ matching elements). As such, in discrete diffusion models, neural networks trained to approximate ratios of the data distribution $\frac{Q_{T-s}(x')}{Q_{T-s}(x)}$ for neighboring $x'$ and $x$ can be used to implement an arbitrarily good approximation to the actual reverse process~\cite{lou2024discretediffusionmodelingestimating}.

\subsection{The Diffusion Loss}

As discussed in the main text, a diffusion model is trained by minimizing the distributional distance between the joint distributions of the forward process $Q_{0, \dots, T}$ and our learned approximation to the reverse process $P_{0, \dots, T}^{\theta}$,
\begin{equation}\label{eq:denoising_loss}
\mathcal{L}_{DN}(\theta) = D\left( Q_{0, \dots, T} (\cdot)|| P^{\theta}_{0, \dots, T}(\cdot)\right)
\end{equation}
the Markovian nature of $Q$ can be taken advantage of to simplify Eq.~\eqref{eq:denoising_loss} into a layerwise form,
\begin{equation}\label{eq:layer_denoising_loss}
\mathcal{L}_{DN}(\theta) + C = -\sum_{t=1}^T \mathbb{E}_{Q(x_{t-1}, x_t)}\left[\log{\left(P_{\theta}(x_{t-1}| x_t \right)}\right]
\end{equation}
where $C$ does not depend on $\theta$. For denoising algorithms that operate in the infinitesimal limit, the simple form of $P_{\theta}$ allows for $\mathcal{L}_{DN}$ and its gradients to be computed exactly.

\subsubsection{A Monte-Carlo gradient estimator}

In the case where $P_{\theta} \left( x^{t-1}|x^t\right)$ is an EBM, there exists no simple closed-form expression for $\nabla_{\theta} \mathcal{L}_{DN}(\theta)$. In that case, one must employ a Monte Carlo estimator to approximate the gradient. This estimator can be derived directly by taking the gradient of Eq.~\eqref{eq:layer_denoising_loss},
\begin{equation}\label{eq:log_p_loss_grad}
\nabla_{\theta} \mathcal{L}_{DN}(\theta) = -\sum_{t=1}^T \mathbb{E}_{Q(x^{t-1}, x^t)} \left[\nabla_{\theta} \log{\left(P_{\theta}(x^{t-1}| x^t ) \right)}\right]
\end{equation}
If we have an EBM parameterization for $P_{\theta}\left( x^{t-1}|x^t\right)$ this may be simplified further. Specifically, given the latent variable from Eq.~8 in the main text, the gradient of log-likelihood may be simplified to,
\begin{equation}
\nabla_{\theta} \log{\left(P_{\theta}(x^{t-1}| x^t ) \right)} = \mathbb{E}_{P_{\theta}(x^{t-1}, z^{t-1}|x^t)}\left[\nabla_{\theta} \mathcal{E}^m_{t-1}\right] - \mathbb{E}_{P_{\theta}(z^{t-1}|x^{t-1}, x^t)} \left[\nabla_{\theta} \mathcal{E}^m_{t-1}\right] 
\end{equation}
Inserting this into Eq.~\eqref{eq:log_p_loss_grad} yields the final result given in Eq.~14 in the article.

\subsection{Simplification of the Energy Landscape}\label{app:simple-energy-landscape}

As the forward process timestep is made smaller, the energy landscape of the EBM-based approximation to the reverse process becomes simpler. A simple 1D example serves as a good demonstration of this concept. Consider the marginal energy function,
\begin{equation}
\mathcal{E}_{t-1}^{\theta} \left( x_{t-1} \right) = \left( x_{t-1}^2 - 1\right)^2
\end{equation}
and a forward process energy function that corresponds to Gaussian diffusion (Eq.~\eqref{eq:cont_diff}),
\begin{equation}
\mathcal{E}_{t-1}^{f} \left( x_{t-1}, x_t \right) = \lambda \left( \frac{x_{t-1}}{x_t} - 1\right)^2
\end{equation}
The parameter $\lambda$ scales inversely with the size of the forward process timestep; that is, $\lim\limits_{\Delta t \to 0} \lambda = \infty$.

\begin{figure}
\includegraphics[width=0.6\linewidth]{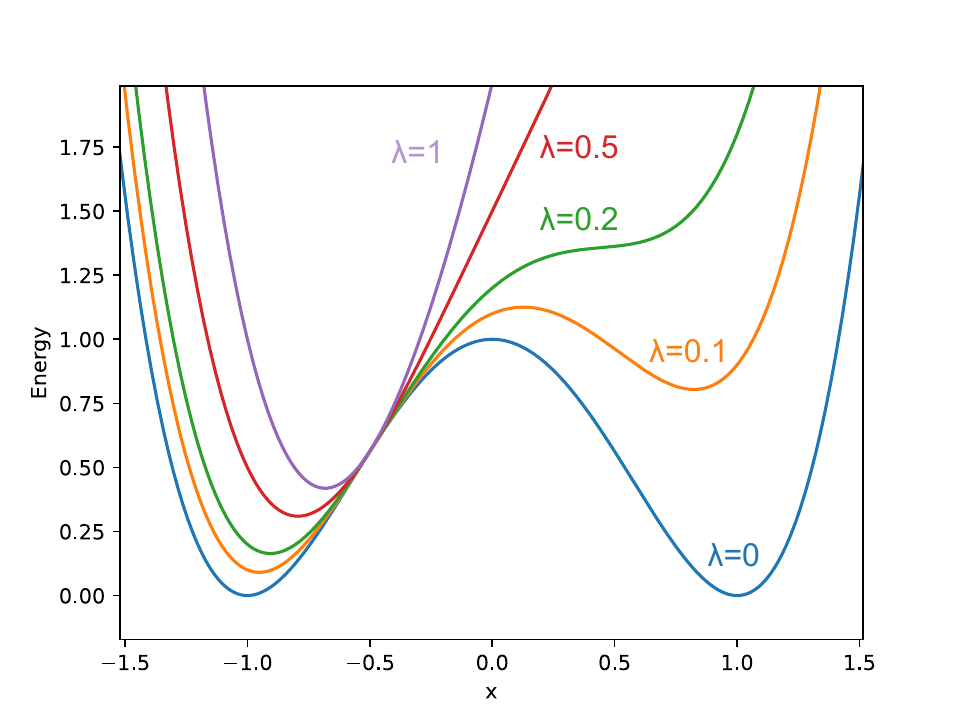}
\caption{\textbf{Conditioning of the energy landscape} As $\lambda$ is increased, the energy landscape is reshaped from a strongly bimodal distribution towards a simple Gaussian centered at $x_{t} = -0.5$. The latter is much easier to sample from.}
\label{fig:energy_conditioning}
\end{figure}

The reverse process conditional energy landscape is then $\mathcal{E}_{t-1}^{\theta} + \mathcal{E}_{t-1}^{f}$. The effect of $\lambda$ on this is shown in Fig.~\ref{fig:energy_conditioning}.

The energy landscape is bimodal at $\lambda = 0$ and gradually becomes distorted towards an unimodal distribution centered at $x_t$ as $\lambda$ increases. This reshaping is intuitive, as shortening the forward process timestep should more strongly constrain $x_{t-1}$ to $x_t$.

\subsection{Conditional Generation}
\label{subsec:conditional_generation}

The denoising framework can be adapted for conditional generation tasks, such as generating MNIST digits given a specific class label. In principle, this is very simple: we concatenate the target (in our case, the images) and a one-hot encoding of the labels into a contiguous binary vector and treat that whole thing as our training data on which we train the denoising model as described above.

In this case, the visible nodes of the Boltzmann machine are partitioned into "pixel nodes" $V_X$ and "label nodes" $V_L$. All visible nodes come in pairs of input and output nodes (drawn in blue and green resp.~in Fig.~3 in the main paper body and Fig.~\ref{fig:hardware_denoising} below), so the set of visible nodes now consists of $\Vin_X, \Vout_X, \Vin_L,$ and $\Vout_L$.

The training procedure works the same way as before, just using this label-augmented data. We obtain the noised training images $X_n$ and labels $L_n$ by noising each entry of $X_0$ and $L_0$ resp.~independently using the forward process described in subsection \ref{subsub:forward_proc_discrete}. Then we train the $n$th step model $P_{\theta_n}\big(\Vout_X = x, \Vout_L = l \,|\, \Vin_X = x' , \Vin_L = l' \big)$ to approximate (in terms Kullback-Leibler divergence) the distribution $\mathbb{P}\big( X_n = x, L_n = l  \,|\, X_{n+1} = x' , L_{n+1}=l' \big)$.

At inference time, we have two cases:
\begin{itemize}
\item Unconditional inference proceeds as with regular denoising. We pass the pixel and label values backward through all the step models, and at the end, we record the pixel values.
\item For conditional generation we clamp all label output nodes $\Vout_L$ in all step models to $l_0$ and sample $\widehat{X}_n \sim P_{\theta_n}\big(\Vout_X = \cdot \,|\, \Vin_X = \widehat{X}_{n+1} , \Vout_L = \Vin_L = l_0 \big)$, where $l_0$ is an unnoised label and $\widehat{X}_{n+1}$ is the output of $P_{\theta_{n+1}}$ (or uniform noise if $n=N$).
\end{itemize}

Note that all step models except $\theta_0$ will be trained on somewhat noised labels, so they might never have seen a pristine unnoised label during training (if there are 10 classes and five label repetitions, a strongly noised label has an approximately $10 \times 2^{-50}$ chance of being a valid unnoised label). However, during conditional inference, the models will have their label nodes clamped to an unnoised label $l_0$, and they may not know how this should influence the generated image (and this problem would only be exacerbated if we clamped to a noised label instead).

This issue can be mitigated by using a rate $\gamma_X$ when noising image entries in the training data and a different rate $\gamma_L$ for noising label entries. Recall that the higher the $\gamma$, the noisier the data will become as $n$ increases.

We consider two extremes:
\begin{itemize}
\item If $\gamma_L \geq \gamma_X$, then we have the exact same problem as before.
\item If $\gamma_L = 0$, then the labels in the training data are a zero-temperature distribution. This low temperature can lead to freezing, potentially negating the benefits denoising could otherwise bring.
\end{itemize}

Experimentally, we observed that settings in the ranges $\gamma_L \in [0.1, 0.3]$ and $\gamma_X \in [0.7, 1.5]$ (for models with four to 12 steps) yielded good conditional generation performance while avoiding the freezing problem.

\subsection{Learning the marginal}\label{app:learning-the-marginal}

If a DTM is trained to match the conditional distribution of the reverse process perfectly, the learned energy function $\mathcal{E}_{t-1}^{\theta}$ is the energy function of the true marginal distribution, that is, $\mathcal{E}_{t-1}^\theta (x) \propto \log Q(x^{t-1})$. To show this, we start by applying the Bayes' rule to the learned reverse process conditional in the limit that it perfectly matches the true reverse process,
\begin{equation}\label{eq:ebm_bayes}
\frac{Q(x^t|x^{t-1}) Q(x^{t-1})}{Q(x^t)} = \frac{1}{Z(\theta, x^t)} e^{-\left( \mathcal{E}^f_{t-1} \left( x^{t-1}, x^t\right) + \mathcal{E}^{\theta}_{t-1}\left(x^{t-1}, \theta\right)\right)}
\end{equation}
defining the distribution,
\begin{equation}\label{eq:learned_marg}
H(x^{t-1}) = \frac{1}{Z(\theta)} \sum_{ z^{t-1}} e^{-\mathcal{E}^{\theta}_{t-1}\left(x^{t-1}, z^{t-1}, \theta\right)}
\end{equation}
\begin{equation}
Z(\theta) = \sum_{x^{t-1}, z^{t-1}} e^{-\mathcal{E}^{\theta}_{t-1}\left(x^{t-1}, z^{t-1}, \theta\right)}
\end{equation}
extracting the forward process from the RHS of Eq.~\eqref{eq:ebm_bayes} and using Eq.~\eqref{eq:learned_marg},
\begin{equation}\label{eq:ebm_frac}
\frac{Q(x^{t-1})}{Q(x^t)} = \frac{Z(\theta) Z}{Z(\theta, x^t)} H(x^{t-1})
\end{equation}

Eq.~\eqref{eq:ebm_frac} can easily be re-arranged into a form where the LHS depends only on $x^t$, and the RHS depends only on $x^{t-1}$. From this, we can deduce,
\begin{equation}
\frac{Q(x^{t-1})}{H(x^{t-1})} = c
\end{equation}
from the fact that $Q$ and $H$ are both normalized, we can find that $c=1$, which establishes the desired equivalence.

\section{Hardware accelerators for EBMs}\label{app:hardwardacc}

In this work, we focus on a hardware architecture for EBMs that are naturally expressed as Probabilistic Graphical Models (PGMs). In a PGM-EBM, the random variables involved in the model map to the nodes of a graph, which are connected by edges that indicate dependence between variables.

PGMs form a natural basis for a hardware architecture because they can be sampled using a modular procedure that respects the graph's structure. Specifically, the state of a PGM can be updated by iteratively stepping through each node of the graph and resampling one variable at a time, using only information about the current node and its immediate neighbors. Therefore, if a PGM is local, sparse, and somewhat heterogeneous, a piece of hardware can be built to efficiently sample from it that involves spatially arraying probabilistic sampling circuits that interact with each other cheaply via short wires.

This local PGM sampler represents a type of compute-in-memory approach, where the state of the sampling program is spatially distributed throughout the array of sampling circuitry. Since the sampling circuits only communicate locally, this type of computer will spend significantly less energy on communication than one built on a Von-Neumann-like architecture, which constantly shuttles data between compute and memory.

Formally, the algorithm that defines this modular sampling procedure for PGMs is called Gibbs sampling. In Gibbs sampling, samples are drawn from the joint distribution $p(x_1, x_2, \dots, x_N)$ by iteratively updating the state of each node conditioned on the current state of its neighbors. For the $i^{th}$ node, this means sampling from the distribution,
\begin{equation}\label{eq:gibbs_update}
x_i[t+1] \sim p(x_i| nb(x_i)[t]).
\end{equation}
This procedure defines a Markov chain whose stationary distribution can be easily controlled by adjusting the conditional update distributions of each node (see the next section for an example). Starting from some random initialization, this iterative update must be applied potentially many times to all graph nodes before the Markov chain converges to the desired stationary distribution, allowing us to draw samples from it.

\begin{figure}
\includegraphics[width=0.6\linewidth]{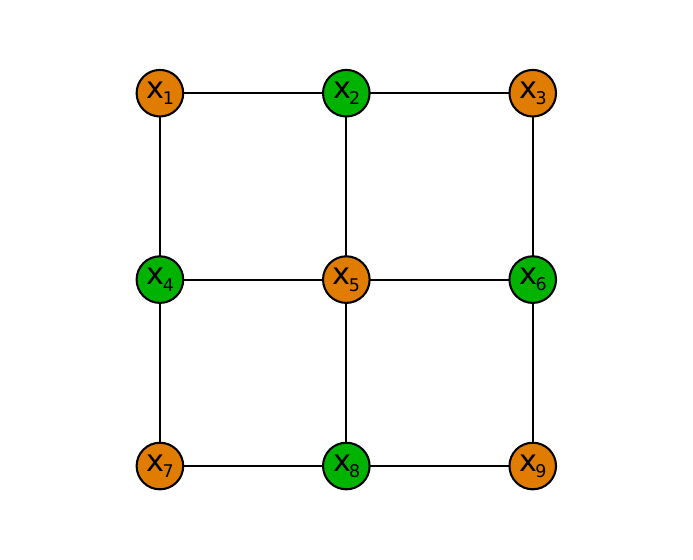}
\caption{\textbf{Chromatic Gibbs Sampling} A schematic view of an abstract hardware accelerator for a simple EBM. Each of the model's variables is assigned to a node. Each node is capable of receiving information from its neighbors and updating its state according to the appropriate conditional distribution. Since each node's update distribution only depends on the state of its neighbors and because nodes of the same color do not neighbor each other, they can all be updated in parallel.}
\label{fig:gibbs_graph}
\end{figure}

Gibbs sampling allows for any two nodes that are not neighbors to be updated in parallel, meaning that the state can be updated in batches corresponding to different color groups of the graph. For a more thorough explanation of how Gibbs sampling works, see ~\cite{pml2Book}.

Fig.~\ref{fig:gibbs_graph} shows a simple example of a PGM with two color groups that would be amenable to Gibbs sampling. Since $x_1$ is only connected to $x_{2}$ and $x_4$, the update rule from Eq.~\eqref{eq:gibbs_update} would take the form,
\begin{equation}
x_1[t+1] \sim p(x_1| x_4[t], x_2[t])
\end{equation}
If the joint distribution had sufficient structure such that the conditional for each node had the same form, a piece of hardware could be built to sample from this PGM by building a 3x3 grid of sampling circuits that communicate only with their immediate neighbors.

\subsection{Quadratic EBMs}

The primary constraint around building a hardware device that implements Gibbs sampling is that the conditional update given in Eq.~\eqref{eq:gibbs_update} must be efficiently implementable. Generally, this means that one wants it to take a form that is "natural" to the hardware substrate being used to build the computer. 

To satisfy this constraint, it is generally necessary to limit the types of joint distributions that a hardware device can sample from. An example of such a restricted family of distributions is quadratic EBMs. 

Quadratic EBMs have energy functions that are quadratic in the model's variables, which generally leads to conditional updates computed by biasing a simple sampling circuit (Bernoulli, categorical, Gaussian, etc.) with the output of a linear function of the neighbor states and the model parameters. These simple interactions are efficient to implement in various types of hardware. As such, Quadratic EBMs have been the focus of most work on hardware accelerators for Gibbs sampling to date. 

In the main text, we discuss Boltzmann machines, which involve only binary random variables and are the most basic form of quadratic EBM. The Conditional Update for Boltzmann Machines requires biasing a Bernoulli random variable according to a sigmoid function of a linear combination of the model parameters and the binary neighbor states, as shown in the main text, Eq.~11. This conditional update is efficiently implementable using an RNG with a sigmoidal bias and resistors, as discussed in section \ref{sec:our_rng_details}.

Here, we will touch on a few other types of quadratic EBM that are more general. Although the experiments in this paper focused on Boltzmann machines, they could be trivially extended to these more expressive classes of distributions.

\subsubsection{Potts models}

Potts models generalize the concept of Boltzmann machines to $k$-state variables. They have the energy function,
\begin{equation}\label{eqn:disc_quadratic_ebm}
E(x) = \sum_{i,j=1}^N \sum_{m,n=1}^M x_m^i J_{mn}^{ij} x_n^j + \sum_{i=1}^N \sum_{m=1}^M h_m^i x_m^i
\end{equation}
\begin{equation}
J^{ii}_{mn} = 0
\end{equation}
$x_m^i$ is a one-hot encoding of the state of variable $x^i$,
\begin{equation}
x_m^i \in \{0, 1 \}
\end{equation}
\begin{equation} 
\sum_m x_m^i = 1
\end{equation}
which implies that $x_m^i = 1$ for a single value of $m$, and is zero otherwise. 
The distribution of any individual variable conditioned on it's Markov blanket is,
\begin{equation}\label{eqn:general_cond}
p(x_m^i=1|\text{mb}(x^i)) = \frac{1}{Z} \exp{\left( -\beta \left( \sum_{j \in \text{mb}(x^i), n} J_{mn}^{ij} x_n^j + \sum_{j \in \text{mb}(x^i), n} x_n^j J_{nm}^{ji} + h_m^i\right)\right)}
\end{equation}
In the case that $J$ has the symmetry,
\begin{equation}
J_{mn}^{ij} = J_{nm}^{ji}
\end{equation}
this reduces to,
\begin{equation}\label{eq:softmax}
p(x_m^i=1|\text{mb}(x^i)) \propto \frac{1}{Z} e^{-\theta_m^i}
\end{equation}

\begin{equation}\label{eq:theta_i}
\theta_m^i = \beta \left( 2 \sum_{j \in \text{mb}(x^i), n} J_{mn}^{ij} x_n^j + h_m^i\right)
\end{equation}
The parameters $\theta$ are defined to make it clear that this is a \textit{softmax} distribution.

Therefore, to build a hardware device that samples from Potts models using Gibbs sampling, one would have to build a softmax sampling circuit parameterized by a linear function of the model weights and neighbor states. Potts model sampling is slightly more complicated than Boltzmann machine sampling, but it is likely possible.

\subsubsection{Gaussian-Bernoulli EBMs}

Gaussian-Bernoulli EBMs extend Boltzmann machines to continuous, binary mixtures. In general, this type of model can have continuous-continuous, binary-binary, and binary-continuous interactions. For simplicity, if we consider only binary-continuous interactions, the energy function may be written as,
\begin{equation}\label{eqn:gb_energy}
E(v,h)=\sum_{i=1}^{N_v}\frac{(v_i-b_i)^{2}}{2\sigma_i^{2}}
       -\sum_{i=1}^{N_v}\sum_{j=1}^{N_h}\frac{v_i W_{ij} h_j}{\sigma_i^{2}}
       -\sum_{j=1}^{N_h} c_j h_j ,
\end{equation}
where $v_i\!\in\!\mathbb{R}$ are continuous variables with biases $b_i$ and variances $\sigma_i^{2}$, $h_j\!\in\!\{-1,1\}$ are binary variables with biases $c_j$, and $W_{ij}$ are interaction weights.

Due to the structure of the energy function, the update rule for the continuous variables corresponds to drawing a sample from a Gaussian distribution with a mean that is a linear function of the neighbor states,
\begin{equation}\label{eqn:gb_cond_v}
p\!\left(v_i\,\middle|\,\text{mb}(v_i)\right)=
\mathcal{N}\!\bigl(\mu_i,\sigma_i^{2}/\beta\bigr),\qquad
\mu_i=b_i+\sigma_i^{2}\sum_{j\in\text{mb}(v_i)} W_{ij} h_j .
\end{equation}

The binary update rule is similar to the rule for Boltzmann machines,
\begin{equation}\label{eqn:gb_cond_h_pm1}
p\!\left(h_j=1\,\middle|\,\text{mb}(h_j)\right)=\sigma\!\left(2\beta\,\left(\sum_{i\in\text{mb}(h_j)} \dfrac{v_i W_{ij}}{\sigma_i^{2}} + c_j\right)\right)
\end{equation}

Hardware implementations of Gaussian-Bernoulli EBMs are more difficult than the strictly discrete models because the signals being passed during conditional sampling of the binary variables are continuous. To pass these continuous values, they must either be embedded into several discrete variables or an analog signaling system must be used. Both of these solutions would incur significant overhead compared to the purely discrete models.

\section{A hardware architecture for denoising}\label{sec:hw}

The denoising models used in this work exclusively modeled distributions of binary variables. The reverse process energy function (Eq.~7 in the main text) was implemented using a Boltzmann machine. The forward process energy function $\mathcal{E}^f_{t-1}$ was implemented using a simple set of pairwise couplings between $x^t$ (blue nodes) and $x^{t-1}$ (green nodes). The marginal energy function $\mathcal{E}^{\theta}_{t-1}$ was implemented using a latent variable model (latent nodes are drawn in orange) with a sparse, local coupling structure. 

\subsection{Implementation of the forward process energy function}\label{app:forward_ebm}

\begin{figure}
\includegraphics[width=\linewidth]{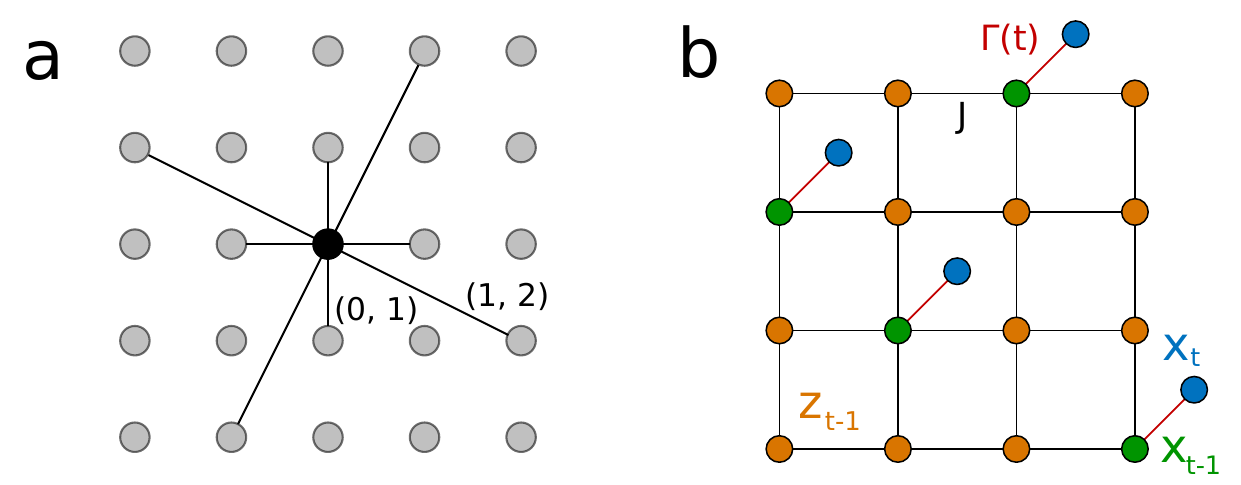}
\caption{\textbf{Our hardware denoising architecture (a)} An example of a possible connectivity pattern as specified in Table.~\ref{tab:graph_edges}. For clarity, the pattern is illustrated as applied to a single cell; however, in reality, the pattern is repeated for every cell in the grid. \textbf{(b)} A graph for hardware denoising. The grid is subdivided at random into visible (green) nodes, representing the variables $x^{t-1}$, and latent (orange) nodes, representing $z^{t-1}$. Each visible node $x^{t-1}_j$ is coupled to a (blue) node carrying the value from the previous step of denoising $x^t_j$ (note that these blue nodes stay fixed throughout the Gibbs sampling). }
\label{fig:hardware_denoising}
\end{figure}

From the exponential form of the discrete-variable forward process transition kernel given in Eq.~\eqref{eq:disc_rev_trans_ker}, it is straightforward to derive a Boltzmann machine-style energy function that implements the forward process,
\begin{equation}\label{eq:forward_energy_app}
\mathcal{E}^{f}_{t-1} = \sum_i \frac{\Gamma_i(t)}{2} x^t_i x^{t-1}_i
\end{equation}
where $x_t[i] \: \in \{-1, 1\}$ indicates the $i^{th}$ element of the vector of random variables $x_t$ as usual. 

\subsection{Implementation of the marginal energy function}\label{sec:marginal}

We use a Boltzmann machine based on a grid graph to implement the marginal energy function. Our grids have both nearest-neighbor and long-range skip connections. A simple example of this is shown in Fig.~\ref{fig:hardware_denoising} (a). This connectivity pattern is tiled such that every node in the bulk of the grid has the same connectivity to its neighbors. At the boundaries, connections that extend beyond the grid's edges are not formed. 

Within the grid, we randomly choose some subset of the nodes to represent the data variables $x_{t-1}$. The remaining nodes then implement the latent variable $z_{t-1}$. The grid is, therefore, a deep Boltzmann machine with a sparse connectivity structure and multiple hidden layers.

We use a particular set of connectivity patterns in the experiments in this article, which are specified in Table.~\ref{tab:graph_edges}. We say that node $(x,y)$ has a connection rule of the form $(a, b)$ if it is connected to nodes at positions $(x+a, y+b), (x-b, y+a), (x-a, y-b), (x+b, y-a)$, so each connection rule adds up to 4 edges from this node.

\begin{table}[]
\centering
\begin{tabular}{||c c||} 
\hline
Pattern & Connectivity \\ [0.5ex] 
\hline\hline
$G_{8}$ & $(0,1),\; (4,1)$ \\ 
\hline
$G_{12}$ & $(0,1),\; (4,1),\; (9,10)$ \\
\hline
$G_{16}$ & $(0,1),\; (4,1),\; (8,7),\; (14,9)$ \\
\hline
$G_{20}$ & $(0,1),\; (4,1),\; (3,6),\; (8,7),\; (14,9)$ \\
\hline
$G_{24}$ & $(0,1),\; (1,2),\; (4,1),\; (3,6),\; (8,7),\; (14,9)$ \\ [1ex] 
\hline
\end{tabular}
\caption{Edges (ordered pairs) associated with graphs of various degrees.}
\label{tab:graph_edges}
\end{table}

As explicitly stated in Eq.~7 of the article, our variational approximation to the reverse process conditional has an energy function that is the sum of the forward process energy function and the marginal energy function. Physically, this corresponds to adding nodes to our grid that implement $x_t$, which are connected pairwise to the data nodes implementing $x_{t-1}$ via the coupling defined in Eq.~\eqref{eq:forward_energy_app}. This connectivity is shown in Fig.~\ref{fig:hardware_denoising} (b).

\section{Energetic analysis of the hardware architecture}\label{app:en-analysis}

Our RNG design uses only transistors and can integrate tightly with other traditional circuit components on a chip to implement a large-scale sampling system. Since there are no exotic components involved that introduce unknown integration barriers, it is straightforward to build a simple physical model to predict how this device utilizes energy.

The performance of the device can be understood by analyzing the unit sampling cell that lives on each node of the PGM implemented by the hardware. The function of this cell is to implement the Boltzmann machine conditional update, as given in Eq.~11 in the main text. 

There are many possible designs for the sampling cell. The design considered here utilizes a linear analog circuit to combine the neighboring states and model weights, producing a control voltage for an RNG. This RNG then produces a random bit that is biased by a sigmoidal function of the control voltage. This updated state is then broadcast back to the neighbors. The cell must also support initialization and readout (get/set state operations). A schematic of a unit cell is shown in Fig.~\ref{fig:gibbs_graph}.

We provide experimental measurements of our novel RNG circuitry in the main text, which establish that random bits can be produced at a rate of $\tau_{rng}^{-1} \approx \: 10 \: \text{MHz}$ using $\sim 350 \text{aJ}$ of energy per bit. Fig.~\ref{fig:rng} (a) shows an output voltage waveform from the RNG circuit. It wanders randomly between high and low states.   Critically, the bias of the RNG circuit (the probability of finding it in the high or low state) is a sigmoidal function of its control voltage, which allows for a straightforward implementation of the conditional update using linear circuitry. 

\begin{figure}
\includegraphics[width=0.6\linewidth]{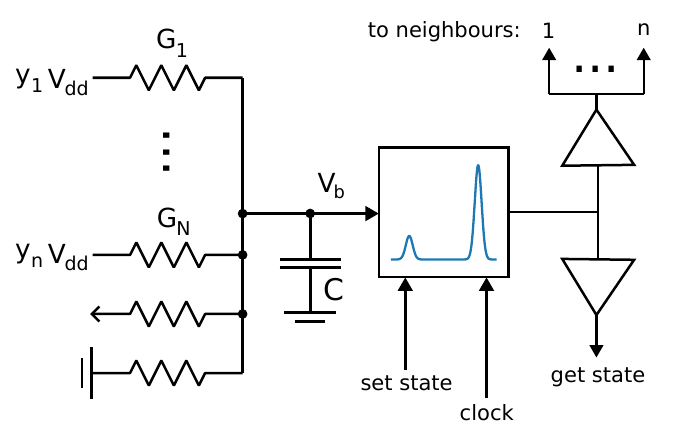}
\caption{\textbf{A schematic of a possible Boltzmann machine sampling cell} A linear resistor network computes a biasing voltage given the sign-corrected neighbor states $y_n = x_n \oplus s_n$. The output of this circuit biases an RNG that responds in a sigmoidal manner. This RNG processes freely when the clock is low and latches to a state when the clock is high. Upon the clock going high, the sampled state is broadcasted to the neighbors of the cell over wires.}
\label{fig:samp_cell}
\end{figure}

The size of the RNG circuit can be used to anchor the dimensions of a future large-scale Gibbs sampling device. As shown in Fig.~\ref{fig:rng} (b), the RNG itself involves around 10 transistors and takes up  $\sim \: 3 \mu m \times 3 \mu m$ on the die. It is reasonable to imagine that the whole sampling cell could fit in $4 \times$ this area and have a side length of $6 \mu m$. Given this area, a $1000 \times 1000$ grid of sampling cells would fit within a $6 \text{mm} \times 6 \text{mm}$ chip.

Building on the measured characteristics of our RNG, we will now develop simple physical models for the remaining components of the sampling system. These models can then be combined to estimate the energy consumption of the diffusion models developed in this article running on our hardware.

\subsection{Biasing circuit}\label{sec:biasing}

The multiply-accumulation of the model weights and neighbor states can be performed using a resistor network, as shown in Fig.~\ref{fig:samp_cell}. The dynamics of this resistor network are described by the differential equation,
\begin{equation}
\sum_{j=1}^{n+2} G_j \left( V_{dd} \: y_j - V_b\right) = C \frac{d V_b}{d t}
\end{equation}
where $y_i = x_i \oplus s_i$ is the XOR of the neighbor state $x_i$ with a sign bit $s_i$. There are $n$ variable neighbor states and two fixed inputs ($y_{n+1} = 1$, $y_{n+2}=0$), which are important for implementing the fixed bias term in the conditional update. $V_{dd}$ is the supply voltage and $V_b$ is the output voltage that biases the RNG. $G_{i}$ represents the conductance of the resistor corresponding to the $i^{th}$ input.  The capacitance $C$ represents the parasitic capacitance to ground associated with any real implementation of this circuit and is critical to forming realistic estimates of speed and energy consumption. Realistic values for an implementation of this circuit in our transistor process are shown in Fig.~\ref{fig:neighbours} (a).

Since this equation is first order, the dynamics exponentially relax to some fixed point $V_{b}^{\infty}$,
\begin{equation}
V_b(t) = c \: e^{-t/\tau_{bias}} + V_{b}^{\infty}
\end{equation}
the time constant $\tau_{bias}$ is,
\begin{equation}
\tau_{bias} = \frac{C}{G_{\Sigma}}
\end{equation}
and the fixed point is,
\begin{equation}\label{eq:bias_fixed_point}
V_{b}^{\infty} = \sum_{j=1}^{n+2} \frac{G_j}{G_{\Sigma}} V_{dd} y_j 
\end{equation}
where the total conductance $G_{\Sigma}$ is,
\begin{equation}
G_{\Sigma} = \sum_{j=1}^{n+2} G_j
\end{equation}

The RNG has a bias curve which takes the form,
\begin{equation}
\mathbb{P}\left( x_i = 1\right) = \sigma \left( \frac{V_b}{V_s} - \phi \right)
\end{equation}
inserting Eq.~\eqref{eq:bias_fixed_point} and expanding the term inside the sigmoid,
\begin{equation}\label{eq:resistor_wb}
\frac{V_b}{V_s} - \phi = \sum_{j=1}^{n} \frac{G_j}{G_{\Sigma}} \frac{V_{dd}}{V_s} \left( x_j \oplus s_j \right) + \left[\frac{G_{n+1}}{G_{\Sigma}} \frac{V_{dd}}{V_s} - \phi \right]
\end{equation}
by comparison to the Boltzmann machine conditional, we can see that the first term implements the model weights (which can be positive or negative given an appropriate setting of the sign bit $s_j$), and the second term implements a bias.

The static power drawn by this circuit can be written in the form,
\begin{equation}
P^{\infty} = \frac{C}{\tau_{bias}} V_{dd}^2 (1-\gamma) \gamma
\end{equation}
where $0 \leq \gamma \leq 1$ is the input-dependent constant,
\begin{equation}
\gamma = \sum_{j=1}^{n+2} \frac{G_j}{G_{\Sigma}} y_j
\end{equation}

This fixed point must be held while the noise generator relaxes, which means that the energetic cost of the biasing circuit is approximately,
\begin{align}\label{eq:bias_energy}
\begin{split}
E_{bias} \approx & P^{\infty} \tau_{rng}  \\
= &  \: C \frac{\tau_{rng}}{\tau_{bias}} V_{dd}^2 (1 - \gamma) \gamma
\end{split}
\end{align}
This is maximized for $\gamma = \frac{1}{2}$.

To avoid slowing down the sampling machine, $\frac{\tau_{rng}}{\tau_{bias}} \gg 1$. As such, ignoring the energy spent charging the capacitor $\sim \frac{1}{2} C V_b^2$ will not significantly affect the results, and the approximation made in Eq.~\eqref{eq:bias_energy} should be accurate. The energy consumed by the bias circuit is primarily due to static power dissipation.

\subsection{Local communication}

Another significant source of energy consumption is the communication of state information between neighboring cells. In most electronic devices, signals are communicated by charging and discharging wires. Charging a wire requires the energy input,
\begin{equation}\label{eq:charge}
E_{\text{charge}} = \frac{1}{2} C_{\text{wire}} V_{\text{sig}}^2
\end{equation}
where $C_{\text{wire}}$ is the capacitance associated with the wire, which grows with its length, and $V_{\text{sig}}$ is the signaling voltage level. 

\begin{figure}
\includegraphics[width=\linewidth]{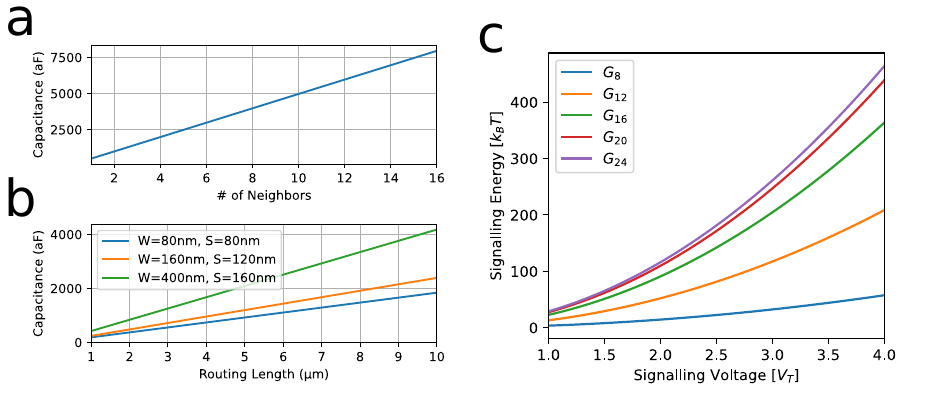}
\caption{\textbf{Parameters for the energy model (a)} The parasitic capacitance associated with the output node of the biasing circuit for various numbers of neighbors. These capacitances were estimated using the PDK and a layout for a real transistor implementation of the biasing circuit. \textbf{(b)} The capacitance associated with routing wires of various lengths and geometry in our process, extracted using the PDK. \textbf{(c)} The energy required for a cell to signal to all of its neighbors as a function of signaling voltage for various connectivity patterns. This energy was calculated using the routing capacitance data from (b).}
\label{fig:neighbours}
\end{figure}

Given the connectivity patterns shown in table ~\ref{tab:graph_edges}, it is possible to estimate the total capacitance $C_n$ associated with the wire connecting a node to all of its neighbors,
\begin{equation}
C_n = 4 \eta \ell \sum_i \sqrt{a_i^2 + b_i^2}
\end{equation}
where $\ell \approx 6 \mu m$ is the sampling cell side length, and $\eta \approx 350 \text{aF}/\mu m$ is the wire capacitance per unit length in our process, see Fig.~\ref{fig:neighbours} (b). $a_i$ and $b_i$ are the $x$ and $y$ components of the $i^{th}$ connection rule, as described in section ~\ref{sec:marginal}.

The charging energy Eq.~\eqref{eq:charge} is plotted as a function of signaling voltage for various connectivity patterns in Fig.~\ref{fig:neighbours} (b).

\subsection{Global communication}

Several systems on the chip require signals to be transmitted from some central location out to the individual sampling cells. This communication involves sending signals over long wires with a large capacitance, which is energetically expensive. Here, the cost of this global communication will be taken into consideration.

\subsubsection{Clocking}

Although it is possible in principle to implement Gibbs sampling completely asynchronously, in practice, it is more efficient to implement standard chromatic Gibbs sampling with a global clock. A global clock requires a signal to be distributed from a central clock circuit to every sampling cell on the chip. This signal distribution is typically accomplished using a clock tree, a branching circuit designed to minimize timing inconsistencies between disparate circuit elements. 

To simplify the analysis, we will consider a simple clock distribution scheme in which the clock is distributed by lines that run the entire length of each row in the grid. The total length of the wires used for clock distribution in this scheme is,
\begin{equation}
L_{clock} = N \: L
\end{equation}
where $N$ is the number of rows in the grid, and $L$ is the length of a row. Given this length, the energetic cost of a clock pulse can be calculated using Eq.~\eqref{eq:charge}.

\subsubsection{Initialization and readout}\label{sec:readout}

A sampling program begins by initializing every sampling cell to a specific state and ends by reading out the state of a subset of the cells for use off-chip. Both of these operations require bits to be sent over a long wire of length $L$ from the chip's boundaries to a sampling cell in the bulk. 

\subsection{Analysis of a complete sampling program}\label{sec:energy_complete}

Given the above analysis of the various subsystems, it is straightforward to construct a model of the energy consumption of a complete denoising model. Running each layer of the denoising model requires initialization of all $N$ nodes, chromatic Gibbs sampling for $K$ iterations, and finally, readout of the $N_{\text{data}}$ data nodes,
\begin{equation}
E =T \left( E_{\text{samp}} + E_{\text{init}} + E_{\text{read}} \right)
\end{equation}

$E_{\text{samp}}$ is the cost associated with the sampling iterations for each layer,
\begin{equation}
E_{\text{samp}} =  K N  \left( E_{\text{rng}} + E_{\text{bias}}+ E_{\text{clock}} + E_{\text{nb}} \right)
\end{equation}
where $E_{\text{clock}}$ and $E_{\text{nb}}$ are the per-cell costs associated with clock distribution and neighbor communication, respectively.

$E_{\text{init}}$ is the cost of initializing all the cells at the beginning of the program,
\begin{equation}
E_{\text{init}} = N \frac{1}{2} \eta L V_{sig}^2
\end{equation}
and $E_{\text{read}}$ is the cost of reading out the data cells at the end,
\begin{equation}
E_{\text{read}} = N_{\text{data}} \frac{1}{2} \eta L V_{sig}^2
\end{equation}

\begin{figure}
\includegraphics[width=\linewidth]{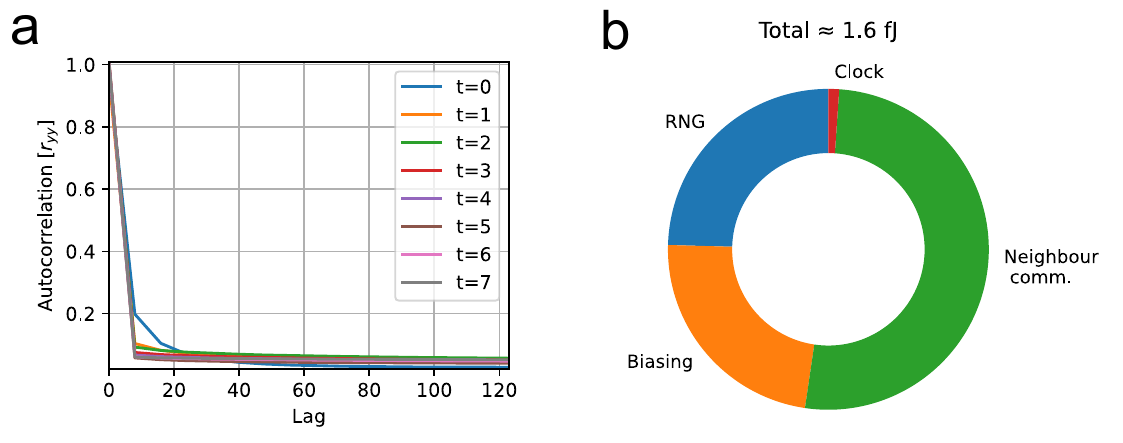}
\caption{(a) \textbf{Autocorrelation curve of a denoising model composed of Boltzmann machines.} Each line represents the autocorrelation of one of the Boltzmann machines that make up a fully trained denoising model. \\
(b) \textbf{Breakdown of the energetic cost of running a sampling cell.} Here, we take $\tau_{rng}/\tau_{bias} = 15$ and $\gamma = 1/2$. We also assume that signaling to neighbors is conducted at a voltage of $ 4 V_T$ (where $V_T$ is the thermal voltage $k_B T/e$) and the clocking and read/write operations are conducted at a signal level of $5 V_T$.} 
\label{fig:cell}
\end{figure}

\begin{figure}
\centering
\includegraphics[width=0.6\linewidth]{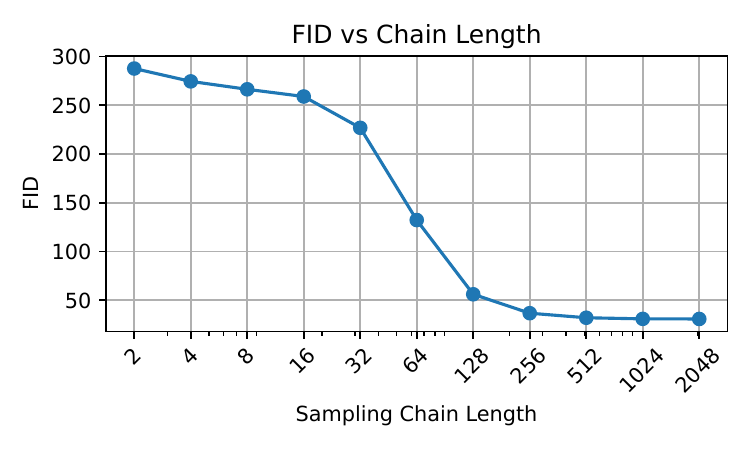}
\caption{The quality of output images generated by our denoising models stops improving when we sample for more than $K \approx 250$ steps.}
\label{fig:fid_v_chain_len}
\end{figure}

This model was used to estimate the energy consumption of the denoising model depicted in Fig. 1 of the article. The mixing behavior for each layer in this denoising model is shown in Fig.~\ref{fig:cell} (a). All of the layers mix in tens of iterations, with the first layer decaying the most slowly. For the sake of energy calculations, we used $K=250$ for all layers to be conservative. Fig.~\ref{fig:fid_v_chain_len} shows that for our trained denoising models, sampling for more than $K\approx250$ steps brings almost no additional benefit, which supports our use of this number for energy calculation. This grid used for each EBM in this model consisted of $N=4900$ nodes that were connected using a $G_{12}$ pattern. $N_{\text{data}} = 834$ of the nodes were assigned to data, and the rest were latent.

Given realistic choices for the rest of the free parameters of the model, the energetic cost of this denoising model is estimated to be around $1.6 \: T \text{nJ}$. This is almost entirely dominated by $E_{\text{samp}}$, with $E_{\text{init}} + E_{\text{read}} \approx 0.01 \: T \text{nJ}$. A breakdown of the various contributions to $E_{\text{samp}}$, along with more details about the used model parameters, is given in Fig.~\ref{fig:cell} (b).

This exact procedure was used to estimate the energy consumption of the MEBMs in Fig. 1 in the article. In this case, $T=1$ and $K$ were estimated from the autocorrelation data for each layer; see section ~\ref{sec:mebm}.

\subsection{Level of realism}

The model presented here captures all of the central functional units of a hardware Boltzmann machine sampler. However, the analysis was performed at a high level, and the model almost certainly underestimates the actual energy consumption of a complete device. In practice, when comparing the results of this type of calculation to a detailed analysis of a complete device design, we generally find agreement within an order of magnitude. Given that the gap between conventional methods and our novel hardware architecture is at least several orders of magnitude, this low-resolution analysis is useful, as it supports the claims made in this article without getting into every implementation detail.

Some of the discrepancies between the high-level and detailed model can be attributed to overheads associated with real circuits. A real implementation of the biasing circuit discussed in section ~\ref{sec:biasing} is more complicated than the theoretical model because tunable resistors do not exist. Communications with neighboring cells over long wires require driver circuits, which consume additional energy beyond what is spent charging the line. Despite this, real circuits are bound by the same fundamental physics as the simplified models presented here. As such, the simplified models tend to estimate energy consumption within a factor of two or three of real-life values.

A real device also has additional supporting circuitry compared to our stripped-down model. In the remainder of this section, we will discuss some examples of such supporting circuitry and argue that their contributions to energy consumption at the system level ought not to be significant. 

\subsubsection{Programming the weights and biases}

Section \ref{sec:biasing} discusses a simple circuit that uses resistors to implement the multiply-accumulate required by the conditional update rule. Key to this is being able to tune the conductance of the resistors to implement specific sets of weights and biases (see Eq.~\eqref{eq:resistor_wb}). 

Practically, implementing this tunability requires that the model parameters be stored in memory somewhere on the chip. Writing to and maintaining these memories costs energy. 

Writing to the memories uses much more energy than maintaining the state. However, if writes are infrequent (program the device once and then run many sampling programs on it before writing again), then the overall cost of the memory is dominated by maintenance. Luckily, most conventional memories are specifically designed to consume as little energy as possible when not being accessed. As such, in practice, the cost of memory maintenance is small compared to the other costs associated with the sampling cells and does not significantly change the outcome shown in Fig.~\ref{fig:cell}.

\subsubsection{Off-chip communication}

External devices have to communicate with our chip for it to be useful. The cost of this communication depends strongly on the tightness of integration between the two systems and is impossible to reason about at an abstract level. As such, the analysis of communication here (as in Section \ref{sec:readout}) was limited to the cost of getting bits out to the edge of our chip, which is a lower bound on the actual cost. 

However, we have found that a more detailed analysis, which includes the cost of communication between two chips mediated by a PCB, does not significantly change the results at the system level. The fundamental reason for this is that sampling programs for complex models run for many iterations before mixing and sending the results back to the outside world. This is reflected in the discrepancy between $E_{\text{samp}}$ and $E_{\text{init}} + E_{\text{read}}$ found in section \ref{sec:energy_complete}.

\subsubsection{Supporting circuitry}

Any real chip has digital and analog supporting circuitry that provides basic functionality, such as clocking and communication, allowing the rest of the chip to function correctly. The fraction of the energy budget spent on this supporting circuitry generally depends on its size compared to the core computer. Due to the heterogeneity of our architecture, it is possible to share most of the supporting circuitry among many sampling cells, which dramatically reduces the per-cell cost. As such, the energy cost of the supporting circuitry is not significant at the system level.

\section{Energy analysis of GPUs}\label{app:energy-gpu}

All experiments shown in Fig. 1 in the article were conducted on NVIDIA A100 GPUs. The empirical estimates of energy were conducted by drawing a batch of samples from the model and measuring the GPU energy consumption and time via Zeus~\cite{you2023zeus}. The theoretical energy estimates were derived by taking the number of model FLOPS (via JAX and PyTorch's internal estimators) and plugging them into the NVIDIA GPU specifications (19.5 TFLOPS for Float32 and 400W). The empirical measurements are compared to theoretical estimates for the VAE in Table \ref{tab:gpu_energy}, and the empirical measurements show good alignment with the theoretical.

\begin{table}[h]
\centering
\begin{tabular}{||c | c | c||} 
\hline
FID & Empirical Efficiency  & Theoretical Efficiency  \\ [0.5ex] 
\hline\hline
30.5 & $6.1 \times 10^{-5}$ & $2.3 \times 10^{-5}$ \\ 
\hline
27.4 & $1.5 \times 10^{-4}$ & $0.4 \times 10^{-4}$ \\ 
\hline
17.9 & $2.5 \times 10^{-3}$ & $1.7 \times 10^{-3}$ \\ 
\hline

\end{tabular}
\caption{Comparing theoretical vs empirical energy consumption for a VAE on a GPU. Energy efficiencies are reported in units of joules per sample.}
\label{tab:gpu_energy}
\end{table}

The models were derived from available implementations and are based on small versions of ResNet~\cite{he2016deep} and UNet~\cite{ronneberger2015u} style architectures. Their FID performance is consistent with published literature values~\cite{dai2018diagnosing, chadebec2022pythae, ostheimer2025sparse}. The goal is not to achieve state of the art performance, but to represent the relative scales of energy consumption of the algorithms. For a direct simulation of the Ising models on a GPU, theoretical efficiency on the order of $10^{-4}$ joules per sample comparable in performance and efficiency to the small VAE.

The reader may be surprised to see that the diffusion model is substantially less energy-efficient than the VAE given the relative dominance in image generation. However, two points should be kept in mind. First, while VAE remains a semi-competitive model for these smaller datasets, this quickly breaks down. On larger datasets, a FID performance gap usually exists between diffusion models and VAEs. Second, these diffusion models (based on the original DDPM~\cite{ho2020denoising}) have performance that can depend on the number of diffusion time steps. So, not only is the UNet model often larger than a VAE decoder, but it also must be run dozens to thousands of times in order to generate a single sample (thus resulting in multiple orders of magnitude more energy required). Modern improvements, such as distillation~\cite{liu2023instaflow}, may move the diffusion model energy efficiency closer to the VAE's.

\section{Autocorrelation and mixing time}
\label{app:autocorr}

We monitor the mixing of our Gibbs samplers via the autocorrelation of a low-dimensional projection of the Markov chain state.

Let $\{x[j]\}_{j \ge 0}$ be a discrete-time Markov chain on a finite state space $\{1, \ldots, d\}$, with time-homogeneous transition kernel $P = (p_{xy})_{1 \le x,y \le d}$ given by
\[
p_{xy} = \mathbb{P}(x[j+1] = y \mid x[j] = x).
\]
In our setting, $x[j]$ is the state of the Boltzmann machine at Gibbs iteration $j$.

We fix a function (projection) $f : \{1, \ldots, d\} \rightarrow \mathbb{R}$ and define the scalar observable
\[
y[j] = f(x[j]).
\]
We then define the mean
\begin{equation}
    \mu \equiv \mathbb{E}\bigl[y[j]\bigr]
    \label{eq:autocorr_mean}
\end{equation}
and the (normalized) autocorrelation function
\begin{equation}
    r_{yy}[k]
    =
    \frac{\mathbb{E}\bigl[(y[j] - \mu)\,(y[j+k] - \mu)\bigr]}{\mathbb{E}\bigl[(y[j] - \mu)^2\bigr]},
    \qquad k \in \mathbb{N}.
    \label{eq:autocorr_def}
\end{equation}
Here $\mathbb{E}[\cdot]$ denotes expectation with respect to the joint law of the Markov chain. In practice, we approximate this expectation by averaging over time and across multiple independent Gibbs sampling chains.

For all experiments in this article, we take $f$ to be the encoder network used in the FID computation, applied to the visible states of the Boltzmann machine. This choice is largely arbitrary: we found that much simpler embeddings, such as random linear projections $y[j] = A x[j]$, behave similarly well for the purposes of autocorrelation-based mixing diagnostics.

\subsection*{Spectral decomposition and decay of autocorrelation}

We now briefly recall how the decay rate of autocorrelation is tied to the spectral properties of the transition kernel $P$, and hence to the mixing time of the Markov chain. Proofs of the statements below can be found in standard references such as~\cite{levin2009markov}.

Assume the Markov chain $\{x[j]\}_{j \ge 0}$ is
\begin{itemize}
\item \textbf{irreducible}, i.e.\ any state can be reached from any other in a finite number of steps, and
\item \textbf{aperiodic}, i.e.\ for any $x \in \{1, \ldots, d\}$ there exists $T \in \mathbb{N}$ such that for all $t \ge T$,
      $\mathbb{P}(x[t] = x \mid x[0] = x) > 0$.
\end{itemize}
Since the chain is finite and irreducible, there exists a unique stationary distribution $\pi$ satisfying
\[
\pi = \pi P.
\]
Furthermore, aperiodicity implies that, for any initial distribution $\psi_0$ of $x[0]$, the law $\psi_t$ of $x[t]$ converges to $\pi$ as $t \to \infty$:
\[
\psi_t = \psi_0 P^t \;\longrightarrow\; \pi
\qquad \text{as } t \rightarrow \infty.
\]

Assume further that $P$ is diagonalizable with real eigenvalues (which is always true if the chain is reversible, like in the case of Gibbs sampling), and write its eigendecomposition as
\[
P = U^{-1} \Sigma\, U,
\]
where $\Sigma$ is diagonal, and its entries are ordered
\[
1 = \sigma_1 > \sigma_2 \ge \cdots \ge \sigma_d \ge 0.
\]
We denote by $U(i,x)$ the entry in the $i$th row and $x$th column of $U$ (and similarly for $U^{-1}$). The first left eigenvector of $P$ (the first row of $U$) is the stationary distribution,
\[
\pi = \pi P = U(1, \cdot),
\]
and the first right eigenvector is a column of ones,
\[
U^{-1}(\cdot, 1) = \mathbf{1}.
\]

Let $f : \{1, \ldots, d\} \to \mathbb{R}$ be the same function as above, and write
\begin{equation}
 \mu^f_0 \equiv \mathbb{E}_{Y \sim \psi_0}\bigl[f(Y)\bigr] = \mathbb{E}\bigl[f(x[0])\bigr],
 \qquad
 \mu^f_{\infty} \equiv \mathbb{E}_{Y \sim \pi}\bigl[f(Y)\bigr]
 = \lim_{t \rightarrow \infty} \mathbb{E}\bigl[f(x[t])\bigr],
 \label{eq:mu0_muinf}
\end{equation}
where $\psi_0$ is the initial distribution of $x[0]$.

Let $\delta_k$ denote the column vector with a $1$ in position $k$ and $0$ elsewhere. Using the eigendecomposition of $P$, the transition probabilities can be written as
\[
(\delta_{x_0}^T P^t)(x)
= \sum_{j=1}^d U^{-1}(x_0, j)\, \sigma_j^t\, U(j, x).
\]
Thus,
\begin{equation}
\begin{split}
 \mathbb{E}\bigl[f(x[0]) f(x[t])\bigr]
 &= \sum_{x_0 = 1}^d f(x_0) \, \mathbb{P}\bigl[x[0] = x_0\bigr]\,
    \mathbb{E}\bigl[f(x[t]) \mid x[0] = x_0\bigr]  \\
 &= \sum_{x_0 = 1}^d \sum_{x=1}^d f(x_0) f(x)\, \psi_0(x_0)\, (\delta_{x_0}^T P^t)(x) \\
 &= \sum_{x_0 = 1}^d \sum_{x=1}^d f(x_0) f(x)\, \psi_0(x_0)\,
    \sum_{j=1}^d U^{-1}(x_0, j)\, \sigma_j^t\, U(j, x) \\
 &= \left( \sum_{x_0 = 1}^d \sum_{x=1}^d f(x_0) f(x)\, \psi_0(x_0)\, U^{-1}(x_0, 1)\, \sigma_1^t\, U(1, x) \right) \\
 &\quad + \sum_{j=2}^d \sigma_j^t
    \sum_{x_0 = 1}^d \sum_{x=1}^d f(x_0) f(x)\, \psi_0(x_0)\, U^{-1}(x_0, j)\, U(j, x).
\end{split}
\label{eq:Ex0xt_expand}
\end{equation}
Using $U^{-1}(\cdot, 1) = \mathbf{1}$ and $U(1, \cdot) = \pi$, the first term simplifies to
\[
\sum_{x_0 = 1}^d \sum_{x=1}^d f(x_0) f(x)\, \psi_0(x_0)\, \pi(x)
= \mu^f_0 \mu^f_\infty.
\]
We can therefore write
\begin{equation}
 \mathbb{E}\bigl[f(x[0]) f(x[t])\bigr]
 = \mu^f_0 \mu^f_\infty + \sum_{j=2}^d \sigma_j^t c_j,
 \label{eq:Ex0xt_spectral}
\end{equation}
where the coefficients $c_j$ are constants (depending on $f$ and $\psi_0$ but not on $t$). For large $t$, the terms with smaller eigenvalues become negligible compared to the contribution from $\sigma_2$, so the covariance
\[
\mathbb{E}\bigl[f(x[0]) f(x[t])\bigr] - \mu^f_0 \mu^f_\infty
\]
decays asymptotically like $\sigma_2^t$. In particular, under stationarity ($\psi_0 = \pi$ so that $\mu^f_0 = \mu^f_\infty = \mu$), the autocorrelation $r_{yy}[t]$ defined in~\eqref{eq:autocorr_def} decays approximately as
\[
r_{yy}[t] \approx C\, \sigma_2^t
\]
for some constant $C$ depending on $f$.

\subsection*{From eigenvalues to mixing time}

The spectral quantity $\sigma_2$ is directly related to the mixing time of the Markov chain. Recall that the mixing time is defined by
\[
\tau(\varepsilon)
= \min\left\{ t \ge 0 : \max_{\psi_0} \|\psi_0 P^t - \pi\|_{\text{TV}} \le \varepsilon \right\},
\]
where
\[
\|\mu - \nu\|_{\text{TV}}
= \frac{1}{2}\sum_{x = 1}^d |\mu(x)-\nu(x)|
\]
denotes the total variation distance and $\varepsilon > 0$ is a prescribed tolerance.

Using the eigendecomposition $P = U^{-1} \Sigma U$, we can write
\[
\psi_0 P^t(x)
=
\sum_{j=1}^d \bigl( \psi_0 \cdot U^{-1}(\cdot, j) \bigr)\, \sigma_j^t\, U(j, x),
\]
where $\psi_0$ is treated as a row vector and $\psi_0 \cdot U^{-1}(\cdot, j)$ denotes the scalar product with the $j$th column of $U^{-1}$. Using $\psi_0 \cdot U^{-1}(\cdot, 1) = 1$ and $U(1, \cdot) = \pi$, we obtain
\begin{align*}
\| \psi_0 P^t - \pi \|_{\text{TV}}
&= \frac{1}{2}\sum_{x = 1}^d \left| \pi(x)
 - \sum_{j=1}^d \bigl( \psi_0 \cdot U^{-1}(\cdot, j) \bigr)\, \sigma_j^t\, U(j, x) \right| \\
&= \frac{1}{2}\sum_{x = 1}^d \left| \pi(x)
 - \pi(x)
 + \sum_{j=2}^d \bigl( \psi_0 \cdot U^{-1}(\cdot, j) \bigr)\, \sigma_j^t\, U(j, x) \right| \\
&\leq \frac{1}{2}\sum_{x = 1}^d \sum_{j=2}^d
 \bigl| \psi_0 \cdot U^{-1}(\cdot, j) \bigr|\,
 |\sigma_j^t|\, |U(j, x)| \\
&= \sum_{j=2}^d \sigma_j^t\, a_j
 \;\leq\; \sigma^t_2 \sum_{j=2}^d a_j,
\end{align*}
where $a_j \ge 0$ are constants that depend on $\psi_0$ and on the eigenvectors of $P$. Therefore, the total variation distance decays at least as fast as $\sigma_2^t$, and we obtain the upper bound
\[
\tau(\varepsilon)
\;\le\;
\frac{\log(\varepsilon) - \log\!\left( \sum_{j=2}^d a_j \right)}{\log(\sigma_2)}.
\]
Since $0 \le \sigma_2 < 1$, the denominator $\log(\sigma_2)$ is negative, and the right-hand side is positive. The smaller the value of $\sigma_2$, the faster the Markov chain mixes, and the more rapidly the autocorrelation $r_{yy}[k]$ decays.

In summary, by empirically estimating the long-lag decay of the autocorrelation function $r_{yy}[k]$ for some informative observable $f(x)$, we effectively probe the second-largest eigenvalue $\sigma_2$ of the transition kernel and thus obtain a proxy for the mixing time of the Gibbs sampler underlying our DTM.

\section{Total correlation penalty}

In the main text (see Eq.~17), we explain how we utilize a total correlation penalty to encourage the latent variable EBMs employed in our model to mix rapidly. Here, we will discuss a few details of this regularizer and the method we use to control its strength adaptively.

\subsection{Gradients of the total correlation penalty}

The total correlation penalty is a convenient choice in this context because its gradients can be computed using the same samples used to estimate the gradient of the usual loss used in training, $\nabla_{\theta} \mathcal{L}_{DN}$. Namely, treating the factorized distribution as a constant with respect to the gradient,
\begin{equation}\label{eq:tc_grad}
\nabla_{\theta} \mathcal{L}^{TC}_t = \mathbb{E}_{Q(x^{t-1})} \left[ \mathbb{E}_{d(s^{t-1}|x^t)}\left[ \nabla_{\theta} \mathcal{E}^{\theta}_{t-1}\right] -\mathbb{E}_{P_{\theta}(s^{t-1}|x^t))}\left[ \nabla_{\theta} \mathcal{E}^{\theta}_{t-1}\right] \right]
\end{equation}
where,
\begin{equation}
d(s^{t-1}|x^t) = \prod_{i=1}^M P_{\theta}(s^{t-1}_i|x^t) 
\end{equation}

The second term in Eq.~\eqref{eq:tc_grad} also appears in the estimator for $ \nabla_{\theta} \mathcal{L}_{DN}$. The first term can be simplified when $\mathcal{E}^{\theta}_{t-1}$ has particular symmetries. For example, if $\mathcal{E}^{\theta}_{t-1}$ is a Boltzmann machine energy function (see main text Eq.~10),
\begin{equation}
\mathbb{E}_{d(s^{t-1}|x^t)}\left[ \frac{d}{d h_i} \mathcal{E}^{\theta}_{t-1}\right] = -\beta \: \mathbb{E}_{P_{\theta}(s_i|x_t)}\left[ s_i\right]
\end{equation}
\begin{equation}
\mathbb{E}_{d(s^{t-1}|x^t)}\left[ \frac{d}{d J_{ij}} \mathcal{E}^{\theta}_{t-1}\right] = -\beta \: \mathbb{E}_{P_{\theta}(s_i|x_t)}\left[ s_i\right] \: \mathbb{E}_{P_{\theta}(s_j|x_t)}\left[ s_j\right]
\end{equation}
Each of these terms is easy to compute given the samples used to estimate $ \nabla_{\theta} \mathcal{L}_{DN}$.

\subsection{Adaptive Correlation Penalty}
\label{app:acp}

The optimal strength of the correlation penalty $\lambda_t$ may vary depending on the specific denoising step $t$ (models for less noisy data near $t=0$ may require stronger regularization) and may even change during training for a single-step model. Manually tuning $\lambda_t$ for each of the step-models would be prohibitively expensive.

To address this, we employ an Adaptive Correlation Penalty (ACP) scheme that dynamically adjusts $\lambda_t$ based on an estimate of the model's current mixing time. We use the autocorrelation of the Gibbs sampling chain, $r_{yy}^t$, as a proxy for mixing, as described in Section \ref{app:autocorr} and the main text, Eq.~18.

Our ACP algorithm monitors the autocorrelation at a lag $K$ equal to the number of Gibbs steps used in the estimation of $ \nabla_{\theta} \mathcal{L}_{DN}$. The goal is to adjust $\gamma_\text{CP}$ to keep this autocorrelation below a predefined target threshold $\varepsilon_\text{ACP}$.

A simple layerwise procedure is used for this control. The inputs to the algorithm are the initial values of $\lambda_t$, a target autocorrelation threshold $\varepsilon_\text{ACP}$ (e.g., $0.03$), an update factor $\delta_\text{ACP}$ (e.g., $0.2$) and a lower limit $\lambda_t^\text{min}$ (e.g., $0.0001$).

At the end of each training epoch $m$:
\begin{enumerate}
\item Estimate the current autocorrelation $a_m^t = r_{yy}^t[K]$. This estimate can be done by running a longer Gibbs chain periodically and calculating the empirical autocorrelation from the samples.
\item Set $\lambda_t' = max(\lambda_t^\text{min}, \lambda_t^{(m)})$ to avoid getting stuck at 0.
\item Update $\lambda_t$ for the next epoch ($m+1$) based on $a_m^t$ and the previous value $a_{m-1}^t$ (if $m>0$):
\begin{itemize}
    \item If $a_m^t < \varepsilon_\text{ACP}$: The chain mixes sufficiently fast; reduce the penalty slightly.
        \[ \lambda_t^{(m+1)} \gets (1 - \delta_\text{ACP}) \lambda_t'\]
    \item Else if $a_m^t \ge \varepsilon_\text{ACP}$ and $a_m^t \le a_{m-1}^t$ (or $m=0$): Mixing is slow but not worsening (or baseline); keep the penalty strength.
        \[ \lambda_t^{(m+1)} \gets \lambda_t' \]
    \item Else ($a_m^t > \varepsilon_\text{ACP}$ and $a_m^t > a_{m-1}^t$): Mixing is slow and worsening; increase the penalty.
        \[ \lambda_t^{(m+1)} \gets (1 + \delta_\text{ACP}) \lambda_t'  \]
\end{itemize}
\item If the proposed value $\lambda_t^{(m+1)} < \lambda_t^\text{min}$, then set $\lambda_t^{(m+1)} \gets 0$.
\end{enumerate}

Our experiments indicate that this simple feedback mechanism works effectively. 
While $\lambda_t$ and the autocorrelation $a_m^t$ might exhibit some damped oscillations for several epochs before stabilizing this automated procedure is vastly more efficient than performing manual hyperparameter searches for $\lambda_t$ for each of the $T$ models.

\begin{figure}
\includegraphics[width=\linewidth]{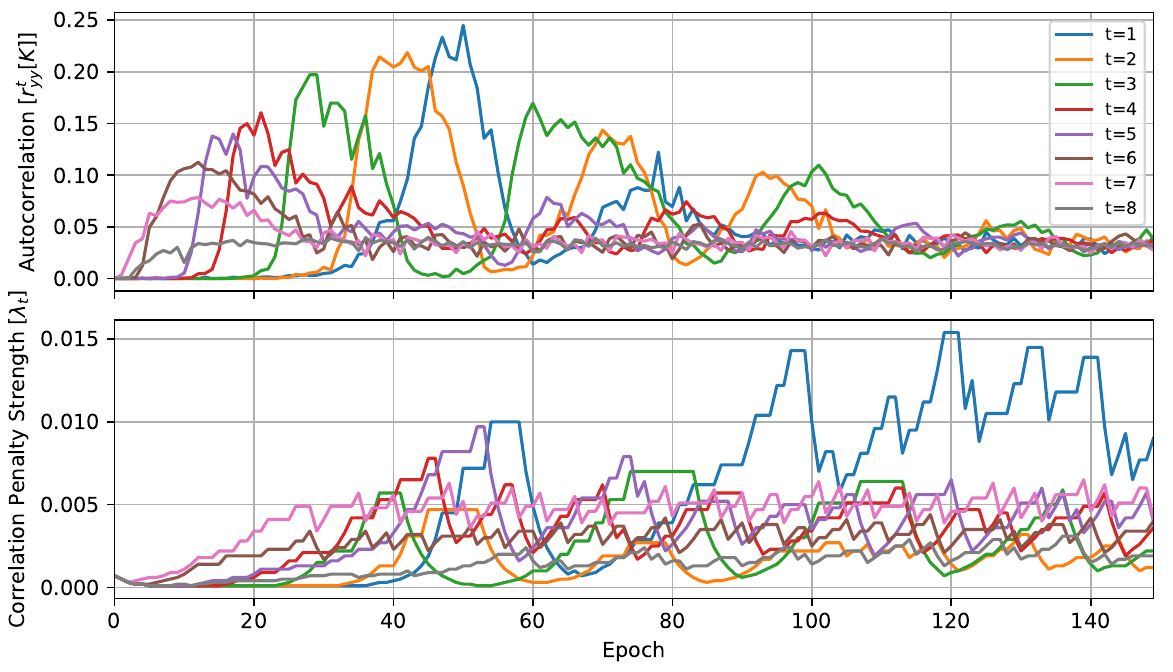}
\caption{\textbf{The adaptive correlation penalty} The dynamics of $r_{yy}^t$ and $\lambda_t$ over a training run. Large values of $r_{yy}^t$ lead to increasingly large values of $\lambda_t$, which cause $r_{yy}^t$ to decrease. The system reaches a stable configuration by the end of training.}
\label{fig:acp_dyn}
\end{figure}

Training is relatively insensitive to the exact choice of $\varepsilon_\text{ACP}$ within a reasonable range (e.g., $[0.02, 0.1]$) and $\delta_\text{ACP}$ (e.g., $[0.1, 0.3]$). Assuming that over the course of training the $\lambda_t$ parameter settles around some value $\lambda_t^*$, one should aim for the lower bound parameter $\lambda_t^\text{min}$ to be smaller than $\frac{1}{2}\lambda_t^*$, while making sure that the ramp-up time $\frac{\log(\lambda_t^*) - \log(\lambda_t^\text{min} )}{\log(1+\delta_\text{ACP})}$ remains small. Settings of $\lambda_t^\text{min}$ in the range $[0.001, 0.00001]$ all produced largely the same result, the only difference being that values on the lower end of that range led to a larger amplitude in oscillations of $\lambda_t$ and $a_m^t$, but training eventually settled for all values. An example of some ACP dynamics is shown in Fig.~\ref{fig:acp_dyn}:

Training on Fashion-MNIST with the typical experimental setup, we observed nearly the same performance (a FID of $28 \pm 1$) for all choices of $\varepsilon_\text{ACP}$, $\delta_\text{ACP}$ and $\lambda_t^\text{min}$ in the ranges written above, so long as we trained for at least 100 epochs (with specific settings the training took longer to converge).

\section{Embedding integers into Boltzmann machines}\label{app:int-embed}

In some of our experiments, we needed to embed continuous data into binary variables. We chose to do this by representing a $k$-state categorical variable $X_i$ using the sum $k$ binary variables $Z_i^k$,
\begin{equation}\label{eq:cat_var_def}
X_i = \sum_{k=1}^{K_i} Z_i^{(k)}
\end{equation}
where $Z_i^{(k)} \in \{0, 1\}$. These binary variables can be trivially converted into spin variables that are $\{-1, 1\}$ valued using a linear change of variables.

Energy functions that involve quadratic interactions between these categorical variables can be reduced to Boltzmann machines with local patches of all-to-all connectivity. For example, consider the energy function,
\begin{equation}
E(x; \theta) = - \sum_{i \neq j} w_{ij} X_i X_j - \sum_{i=1}^d b_i X_i
\end{equation}
inserting Eq.~\eqref{eq:cat_var_def}, we can rewrite this in terms of quadratic interactions between the underlying spins $Z_i^{(k)}$,
\begin{equation}
E(z; \theta) = - \sum_{i \neq j} w_{ij} \left( \sum_{k=1}^{K_i} Z_i^{(k)} \right) \left( \sum_{l=1}^{K_j} Z_j^{(l)} \right) - \sum_i b_i \left( \sum_{k=1}^{K_i} Z_i^{(k)} \right)
\end{equation}
which is a standard Boltzmann machine energy function that can be run on our hardware, just like any other.

\section{Deterministic embeddings for DTMs}\label{app:determ-embed}

In \cref{sec:htdml} of the paper, we mention hybrid thermodynamic models, the purpose of which is to combine the flexibility of conventional neural networks (NNs) with the efficiency of probabilistic computers. For example, in the context of image generation, a small convolutional neural network can be used to map color images into a format compatible with a binary DTM. To properly take advantage of the DTM's energy efficiency, the conventional model should be at least 2 or 3 orders of magnitude smaller (e.g., in terms of parameter count or number of operations per sample) than the DTM.

There are various options for the type of conventional model one can use for the embedding, e.g., invertible models such as GLOW~\cite{kingma2018Glow} or Normalizing Flows~\cite{Papamakarios2021NF}, as well as simpler solutions, such as an Autoencoder.

For our proof-of-concept for hybrid models, we used a combination of an Autoencoder and a GAN~\cite{Goodfellow2014GAN}.
\begin{itemize}
\item First, we train a convolutional Autoencoder (encoder plus decoder) that maps images into a binary latent space (achieved through a combination of a sigmoid activation, a binarization penalty, and a straight-through gradient).
\item Second, we train a DTM on latent embeddings of the training images. At inference time, the samples generated by the DTM are passed through the decoder to produce images.
\item Thirdly, we use a GAN-like approach to fine-tune the decoder to utilize the outputs of the Boltzmann machine maximally. Specifically, the Boltzmann machine outputs are used as the noise source, which is fed into the decoder (now taking the role of the generator in the GAN architecture), and finally, a critic is trained to guide the decoder towards generating higher-quality images.
\end{itemize}

Our hybrid model achieved a FID score of $\sim 60$ on CIFAR10. Our DTM had 8 million parameters, the decoder had 65k, and the encoder and critic were both below 500k parameters. At inference time, only the DTM and the decoder are used. To achieve a similar performance with a conventional GAN, the decoder/generator requires about 500000 parameters. 

\section{Some details on our RNG}
\label{sec:our_rng_details}

\begin{figure}
\includegraphics[width=\linewidth]{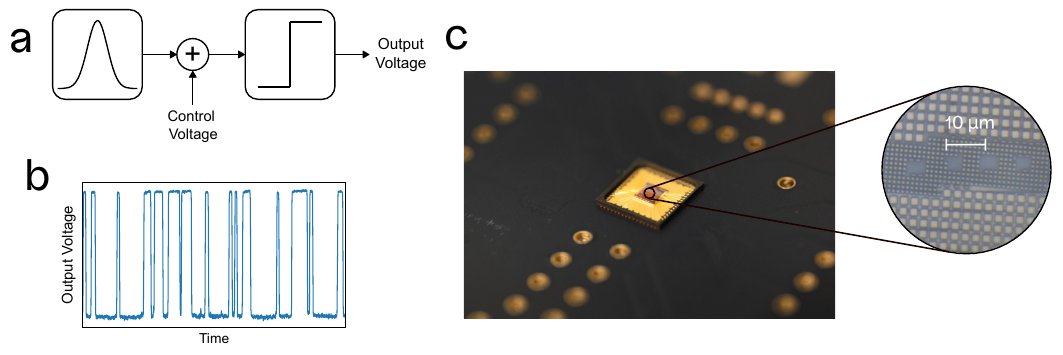}
\caption{\textbf{Our RNG}. \textbf{(a)} A high-level schematic of our RNG design. \textbf{(b)} Stochastic voltage signal from our RNG. The high level represents one, and the low level represents 0. The signal wanders randomly between high and low levels, with the amount of time it spends in each level controlled by the bias voltage. \textbf{(c)} An image of our packaged test chip (with the top of the package removed) assembled onto a PCB. We also show an optical microscope image of several RNG circuits on our test chip. Each circuit occupies an approximately $3 \times 3 \mu m$ area on the chip.} 
\label{fig:rng}
\end{figure}

Our RNG is a digitizing comparator fed by a source of Gaussian noise. The noise source is implemented using the circuitry and principles described in~\cite{our_gyrator}. The comparator is a standard design that operates in subthreshold to minimize energy consumption. The mean of the Gaussian noise is shifted before it is sent into the comparator to implement the bias control. A schematic of our RNG is shown in Fig.~\ref{fig:rng} (a).

Another example of an output voltage signal from our RNG is shown in Fig.~\ref{fig:rng} (b). The signal randomly wanders between high and low-voltage states. Suppose this signal is repeatedly observed, waiting for at least the correlation time of the circuit between observations. In that case, one will approximately draw samples from a Bernoulli distribution with a bias parameter that depends on the circuit's control voltage.

Our RNG was part of the same test chip used to carry out the experiments in~\cite{our_gyrator}. The output of the RNG was fed into an amplification chain that buffered it and allowed its signal to be observed using an external oscilloscope. Fig.~\ref{fig:rng} (c) shows an image of our packaged test chip, along with a view of our RNG through a $100\times$ microscope objective.

\section{MEBM experiments}\label{sec:mebm}

Our experiments on MEBMs were conducted in the typical way~\cite{kerem_sparse}. We employed the same Boltzmann machine architecture as we typically use for the DTM layers, specifically $L=70$ with $G_{12}$ connectivity. Random nodes were chosen to represent the data, and the rest were left as latent variables (as discussed in section \ref{sec:hw}).

Generating the data presented in Figs. 1 and 2 in the main text required controlling the mixing time of a trained Boltzmann machine. To achieve this, we added a fixed correlation penalty (Eq.~17 in the main text) and varied the strength to control the allowed complexity of the energy landscape. 

\begin{figure}[H]
\includegraphics[width=\linewidth]{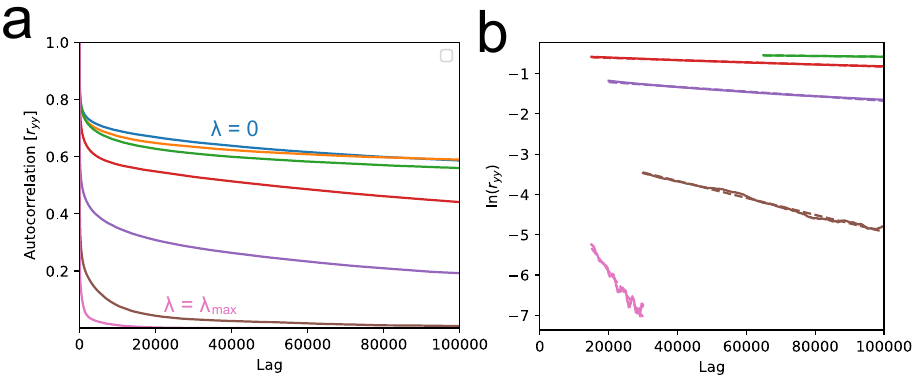}
\caption{\textbf{Boltzmann machine autocorrelation curves (a)} The raw autocorrelation data associated with Boltzmann machines trained using different values of the parameter $\lambda$. \textbf{(b)} The log of the long-time autocorrelation for some of the curves shown in (a). All curves, except for the blue and orange ones, eventually became linear.} 
\label{fig:bm_ryy}
\end{figure}

Fig.~\ref{fig:bm_ryy} (a) shows an example of the raw autocorrelation curves produced by sampling from Boltzmann machines trained with different correlation penalty strengths. The slowest exponential decay rate ($\sigma_2$) could be estimated for most of the curves by fitting a line to the natural log of the autocorrelation curve at long times, see Fig.~\ref{fig:bm_ryy} (b) The two curves with the smallest correlation penalty did not reduce to simple exponential decay during the measured lag values, which means the decay rate was too long to be extracted from our data.

The exponential decay rates extracted from Fig.~\ref{fig:bm_ryy} were used as the mixing times in Fig. 2 in the article. Calling this a "mixing time" is a slight abuse of nomenclature. However, we did not think it made enough of a difference to the article's message to disambiguate (since it is an upper bound on the mixing time, as discussed in Section \ref{app:autocorr}). 

\section{Sampling time vs Performance Tradeoff}
\label{app:sampling_time_and_acp}
The total time it takes to sample from a DTM is $O(KT)$, where $T$ is the number of denoising steps comprising the DTM and $K$. While one might use a different number of Gibbs steps at training time and inference time, it should be noted that for DTMs trained with adaptive correlation penalty (ACP), $K_\text{inference}$ should be similar or only slightly higher to $K_\text{train}$. This is substantially different from the usual good practice for un-penalised EBMs. The main reason for this distinction is ACP. In particular, when training an EBM (or DTM) with ACP, the closed-loop control used in ACP will make sure that the mixing time $K_\text{mix}$ of the EBM is approximately equal to $K_\text{train}$. It is always the case that using $ K_\text{inference} >> K_\text{mix}$ brings no benefits compared to $K_\text{inference} \approx K_\text{mix}$ and therefore when sampling from an ACP-regularised model, using $K_\text{inference} \approx K_\text{train}$. On the other hand, for usual EBMs trained without ACP (or a similar penalty), the mixing time of the EBM can greatly exceed $K_\text{train}$. Due to the mixing-expressivity tradeoff, this at first means that the performance of this un-penalised EBM can exceed that of an ACP EBM, as long as $K_\text{inference} \geq K_\text{mix} >> K_\text{train}$. However, if we then keep training with $K_\text{train} << K_\text{mix}$, the performance of the EBM starts to degrade substantially, as shown in \cref{fig:mix-exp_training}. One may attempt to circumvent this by treating the EBM as a non-equilibrium model as explored in \cite{Decelle_2022}, but that introduces significant new complexity and is mathematically not well understood.

That being said, why does increasing $K_\text{train}$ improve the performance of an ACP-regularised DTM or MEBM? ACP works by periodically checking whether $K_\text{mix} > K_\text{train}$. If yes, then it penalizes the model more, reducing $K_\text{mix}$, but potentially hurting the expressivity of the model. If not, then it reduces the penalty coefficient, allowing both $K_\text{mix}$ and model expressivity to increase. Therefore, using a larger $K_\text{train}$ lets ACP be more lenient, resulting in a more expressive model.

\begin{figure}[H]
    \centering
    \includegraphics[width=0.6\linewidth]{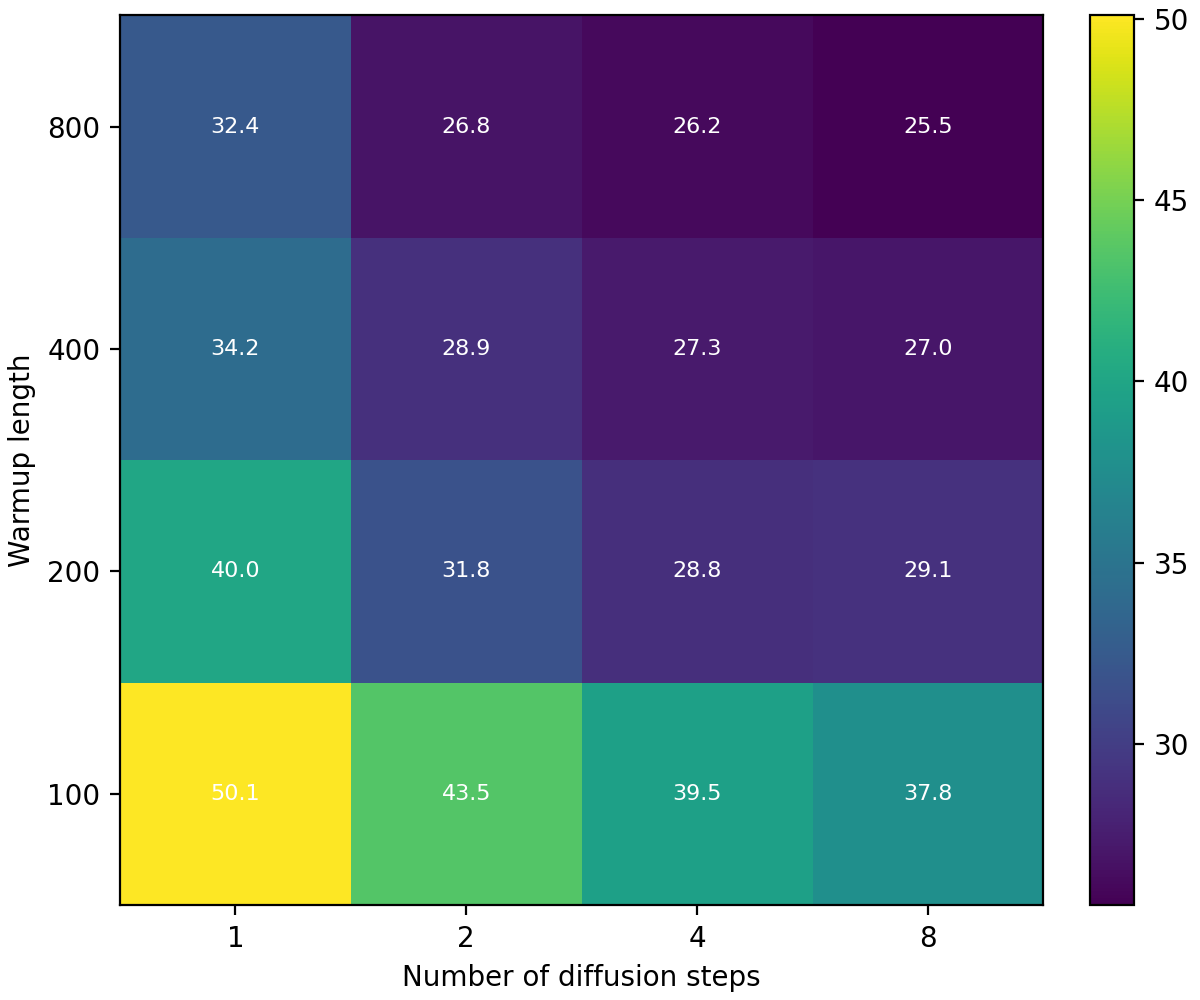}
    \caption{The FID scores of different DTMs trained on binarised FashionMNIST. On the horizontal axis we vary the number of denoising steps $T$ and on the vertical axis we vary the number of Gibbs steps during training $K_\text{train}$. Each diagonal (running from top-left to bottom-right) represents a constant amount of energy expenditure. For this plot, the number of Gibbs steps at inference was twice the number used during training. The reported FID values were averaged over three runs with different random seeds.}
    \label{fig:steps_cd_fid_phase_diagram}
\end{figure}

\begin{figure}
    \centering
    \includegraphics[width=0.5\linewidth]{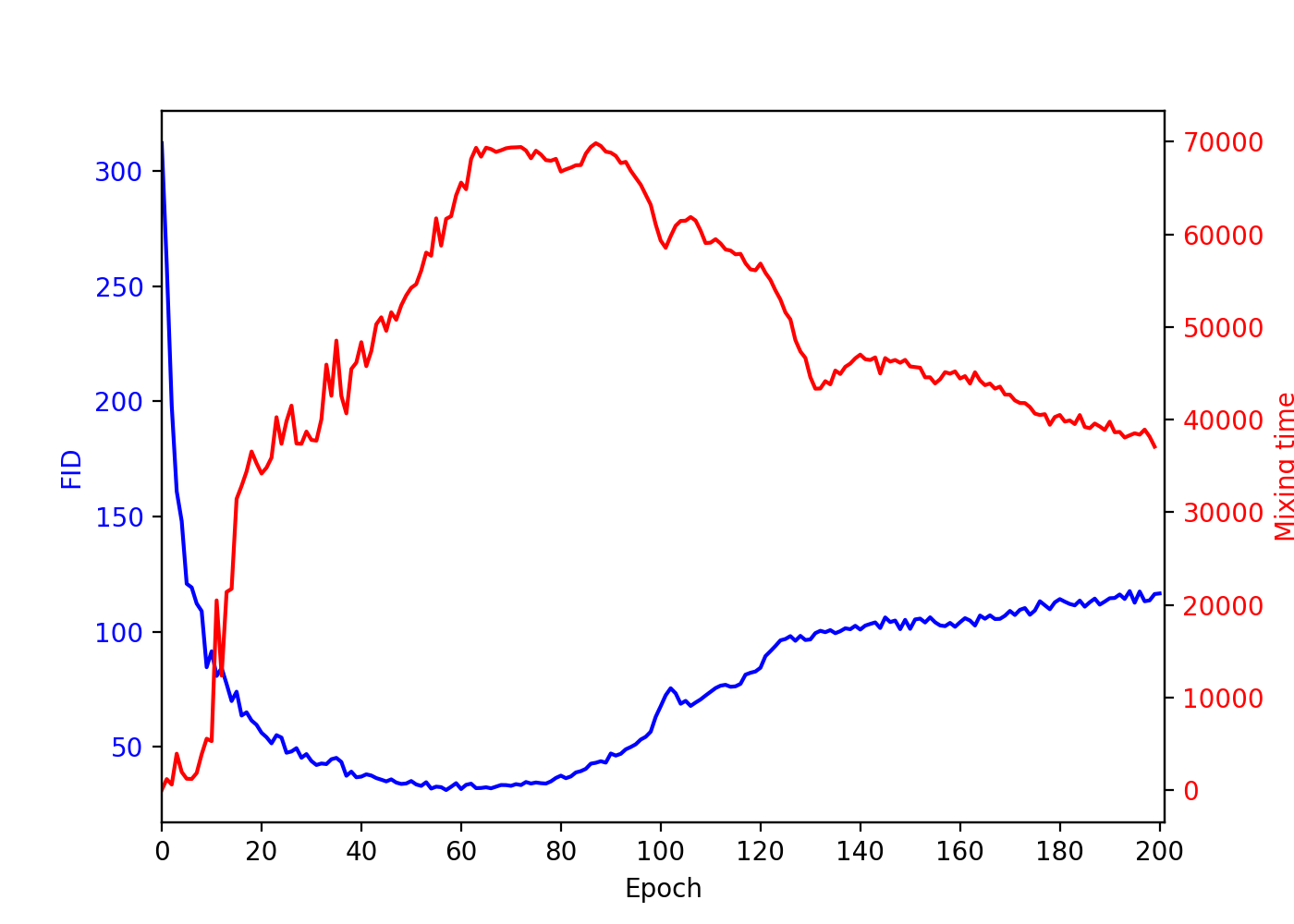}
    \caption{FID and mixing time of an MEBM (same topology as most other experiments in this article) throughout training. As the MEBM trains and becomes more expressive (FID goes down) the mixing time increases. Around epoch 20 the model starts freezing (the mixing time becomes too long), which causes the CD gradients to become inaccurate and soon after the performance of the model starts to degrade. The mixing time subsequently also decreases somewhat, although this is likely an artefact of mode collapse rather than a sign of relaxation of the energy barriers in the model.}
    \label{fig:mix-exp_training}
\end{figure}

\section{CIFAR-10 Images}
\label{app:cifar_imgs}
\begin{figure}[H]
    \centering
    \includegraphics[width=0.9\linewidth]{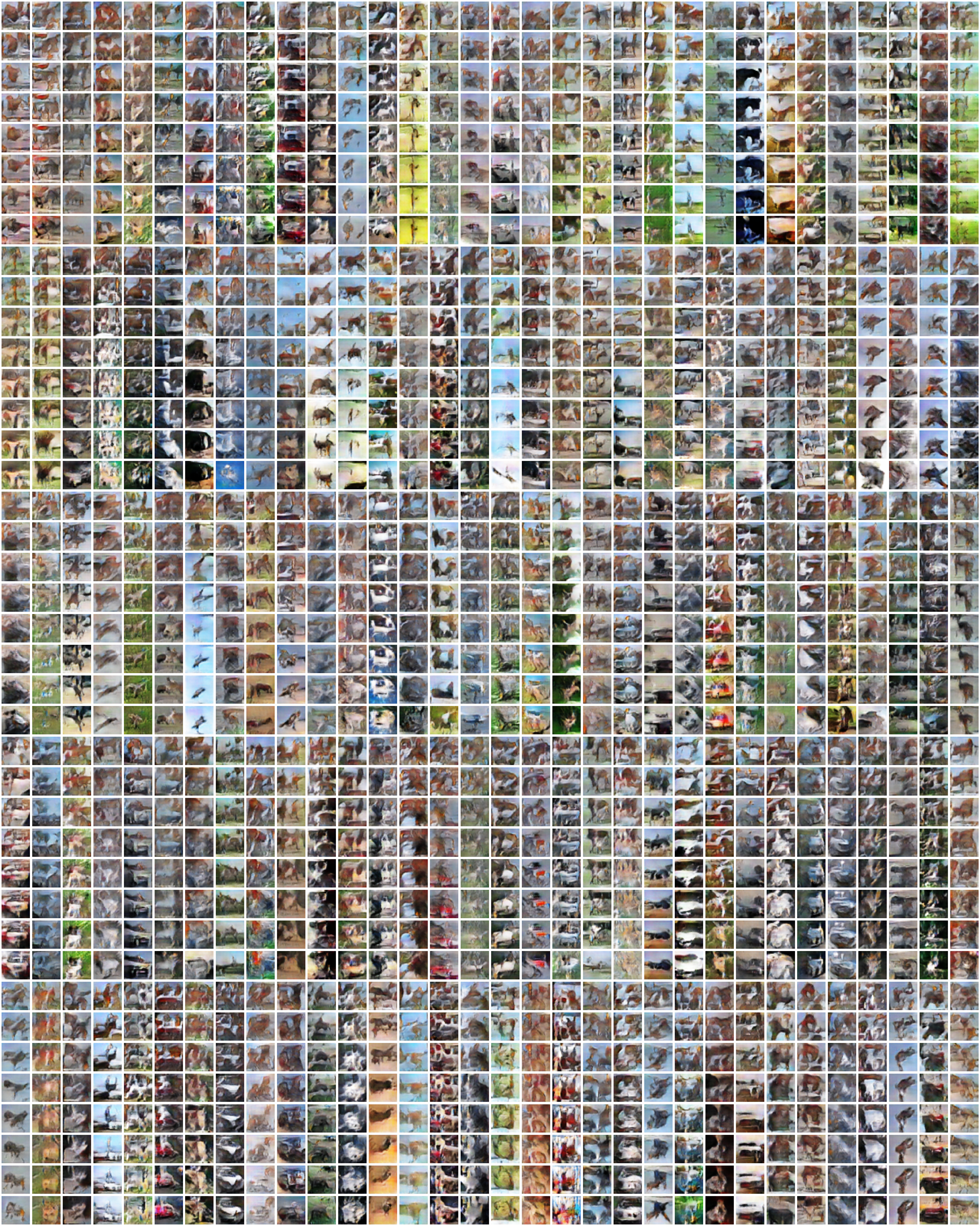}
    \caption{Images generated by the hybrid model trained on CIFAR-10 as described in \cref{sec:htdml} and Appendix~\ref{app:determ-embed}. Images come in groups of 8, where each group shows the progression of the image throughout the denoising process. The denoising process operates inside a binary latent space, which is then converted into images by a small neural-network-based decoder.}
    \label{fig:cifar_imgs}
\end{figure}

\newpage

\putbib[settings,sup]  
\end{bibunit}

\end{document}